\documentclass{article}
\PassOptionsToPackage{numbers,compress}{natbib}

\usepackage[preprint]{neurips_2026}

\usepackage[utf8]{inputenc} 
\usepackage[T1]{fontenc}    
\usepackage{hyperref}       
\usepackage{url}            
\usepackage{booktabs}       
\usepackage{amsfonts}       
\usepackage{nicefrac}       
\usepackage{microtype}      
\usepackage{xcolor}         
\usepackage{natbib}
\usepackage{subcaption}
\usepackage{xspace}
\newcommand{\ie}{i.e.}
\newcommand{\eg}{e.g.}
\usepackage{times}
\usepackage{epsfig}
\usepackage{graphicx}
\usepackage{amsmath}
\usepackage{amssymb}
\newcommand{\E}{\mathbb{E}}
\DeclareMathOperator{\KL}{KL}
\usepackage{algorithm}
\usepackage[noend]{algpseudocode}

\usepackage{color}
\usepackage{booktabs}
\usepackage{graphicx}
\usepackage{hyperref}     
\usepackage[capitalise]{cleveref}     
\usepackage{amsthm}
\theoremstyle{definition}
\newtheorem{assumption}{Assumption}[section]
\newtheorem{proposition}{Proposition}[section]
\usepackage{wrapfig}
\title{Balancing Image Compression and Generation with Bootstrapped Tokenization}

\author{%
  \textbf{Haozhe Chi} $^1$\; \textbf{Jinghan Li} $^1$\;\textbf{Hao Jiang} $^1$\; \textbf{Wu Sheng} $^1$\;\textbf{Yi Ma} $^2$\; \textbf{Jing Wang} $^2$\; \textbf{Yadong Mu} $^1$\thanks{Corresponding author.} \\
  $^1$Peking University, $^2$Central Media Technology Institute, Huawei
}

\begin{document}

\maketitle

\begin{abstract}
  Despite progress in image tokenization, standard methods encode redundant information by mixing all granularities within each token, thus redundancy persists between tokens. The mix of information of different granularity also complicates the training of generators. This paper introduces SelfBootTok, a method that resolves this by cleanly decomposing information into global and local token groups. Through self-bootstrapped learning, the model predicts local details exclusively from global tokens, shifting the burden of visual details from the generator to the tokenizer. Consequently, our generator is far more efficient, requiring only global tokens and reducing computation by approximately 40\%, while delivering superior reconstruction and generation. Moreover, this paradigm scales elegantly: by leveraging more data or parameters to self-supervise local representation learning, SelfBootTok achieves a new state-of-the-art gFID score of 1.56 using only 64 tokens.
\end{abstract}

\section{Introduction}
Recent advancements in diffusion-based image generation, exemplified by the Diffusion Transformer (DiT)~\cite{peebles2023scalable} and flow matching~\cite{lipman2022flow, gat2024discrete, dao2023flow}, have driven significant progress. Beyond diffusion models, other generative paradigms like masked generative models and autoregressive visual models have seen notable developments. For instance, VAR~\cite{tian2024visual} introduces next-scale prediction, where autoregressive visual models surpass diffusion models. Additionally, MAE-tok~\cite{chen2025masked} and MAR~\cite{li2024autoregressive} have improved masked generative modeling, boosting downstream generation performance. Multi-scale designs, as demonstrated by HieraTok~\cite{chen2025hieratok}, VAR~\cite{tian2024visual}, and FlowAR~\cite{ren2024flowar}, further enhance performance.
In parallel with quality improvements, recent studies have focused on enhancing generation efficiency. For example, Lightning-DiT~\cite{yao2025reconstruction} accelerates training by optimizing latent representations for image generation, addressing the reconstruction-generation tradeoff. Similarly, REG~\cite{wu2025representation} accelerates SiT~\cite{ma2024sitexploringflowdiffusionbased} training. As generative frameworks advance, image tokenization has become essential for enabling multimodal understanding and generation through compact latent representations. Early methods like VQ-GAN~\cite{esser2021taming} and VQ-VAE~\cite{van2017neural} encoded images as 2D grid latents, preserving spatial relationships. However, the need for a strict one-to-one mapping between image patches and tokens is unclear, driving the exploration of 1D sequential tokenizers, which offer higher compression while retaining key semantic and structural information.


\begin{figure}[t!]
    \centering
    \includegraphics[width=0.7\linewidth]{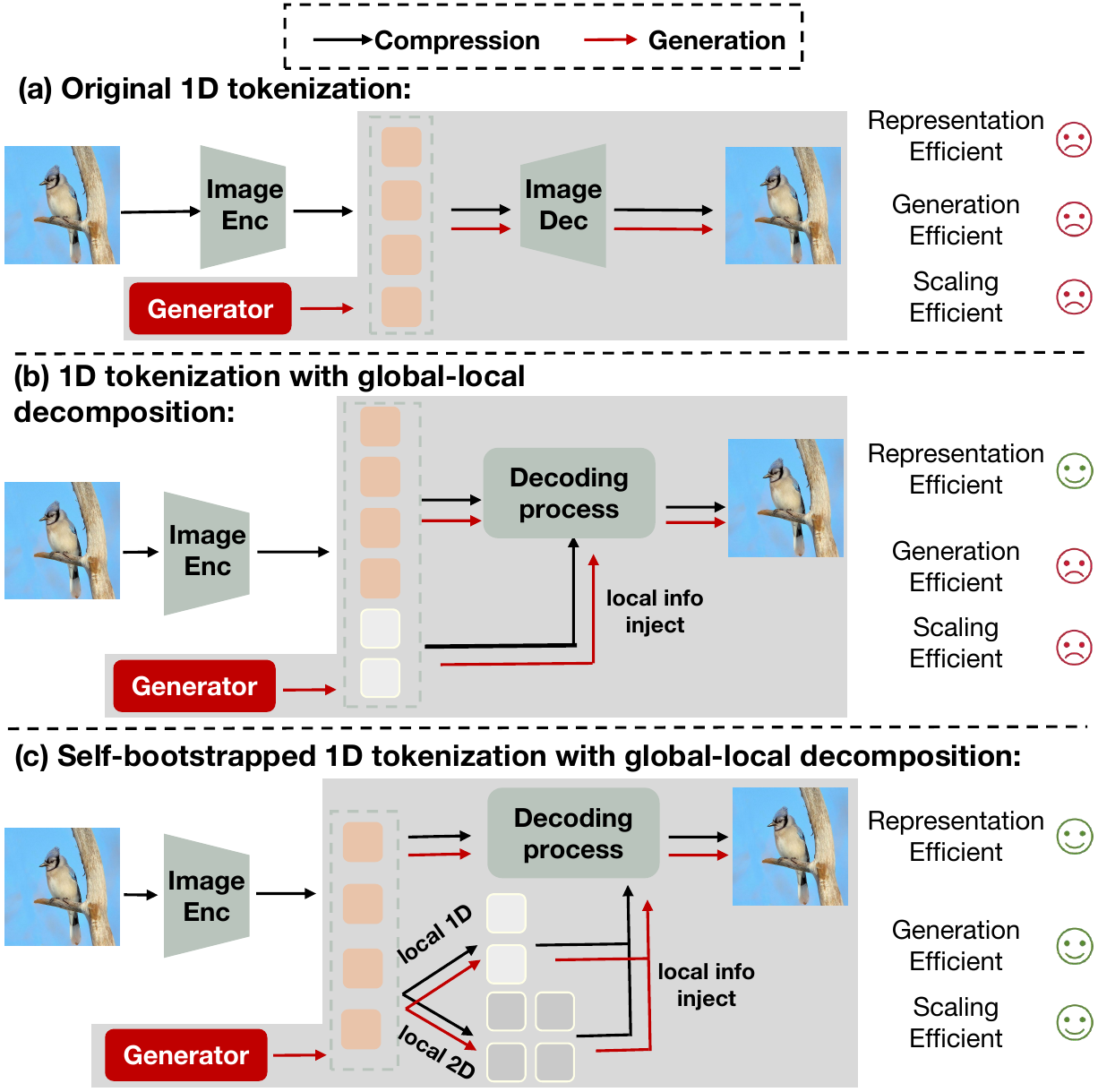}
    \vspace{-0.1in}
    \caption{Illustration of our self-bootstrapped learning paradigm. Compared to classical 1D image tokenizers~\cite{yu2024image, chen2025softvq, xiong2025gigatok} and recent approaches incorporating local detail injection~\cite{chen2025hieratok, esteves2025spectral}, our method employs a global-local decomposition to achieve efficient hierarchical representation learning and adopts a self-bootstrapped strategy to enable both efficient generation and scalable training.}
    \label{fig:architectureorig}
\vspace{-0.2in}
\end{figure}


The emergence of 1D image tokenizers to encode images into compact latent sequences provides a promising alternative to conventional 2D grid representations. The key challenge lies in capturing both global semantics and local details within an extremely limited number of tokens. Early efforts such as Titok~\cite{tian2024visual} demonstrated that an image can be effectively represented by as few as 32 one-dimensional tokens, establishing the feasibility of this paradigm. Subsequent developments, including Flextok~\cite{bachmann2025flextok}, TA-Titok~\cite{kim2025democratizing}, Flowtok~\cite{he2025flowtok}, and GigaTok~\cite{xiong2025gigatok}, have progressively enhanced token efficiency, flexibility, and semantic richness.
Despite these advancements, existing methods still exhibit redundant token interactions and overlap between global and local representations, which constrain scalability and generation efficiency.

In this work, we propose SelfBootTok. It improves the balance at various aspects between image compression and generation, by decomposing conventional visual tokens into high-level semantics-based (\ie, global tokens) and fine-grained visual information-related (\ie, local tokens). The conceptual comparison with existing works is illustrated in~\cref{fig:architectureorig}. The proposed new model has several advantages:

\textbf{1. Simpler and more efficient generator}: Traditionally, visual generators are required to produce all levels of visual details at once when fed with prompts. In contrast, our method bootstraps fine-grained visual information directly from global tokens using unlabeled images, as shown in ~\cref{fig:architectureorig}, thereby bypassing the need for extensive text-image pairs. This scheme essentially moves part of the generation process into the compression pipeline. The generator can focus solely on learning high-level semantics from a compact token set, requiring significantly less data and complexity.

\textbf{2. Scalable image tokenizer}: Our tokenizer employs a self-supervised learning strategy to reconstruct local image information from global tokens. It utilizes a hybrid of 1D and 2D local tokens to capture features at varying granularities. A novel optimal transport alignment is introduced to compactly map these 2D features into the 1D token sequence. This paradigm allows the tokenizer to scale efficiently using more data or parameters. This decomposition also minimizes token-level redundancy, thereby reducing the computational burden on the subsequent generator and enhancing the generation efficiency.

\textbf{3. Parallel optimization of scaled-up tokenizers and generators}: This framework enables efficient scaling by sharing only a set of global tokens. After learning these tokens, the global component of the tokenizer is frozen. The generators and the larger, bootstrapped local components of the tokenizers can then be trained in parallel. Crucially, this decouples their design, allowing the tokenizer size to be varied without requiring the generators to be re-trained.


Comprehensive experiments demonstrate that our method achieves state-of-the-art generation performance among 1D tokenizers and exhibits strong scalability of the self-bootstrapping design. Moreover, we propose a training strategy that scales local aligners while generating global tokens only once, reducing total computational cost by approximately 40\% and training time by about 54\%.

\section{Related work}
\begin{figure*}[th!]
    \centering
    \includegraphics[width=0.95\linewidth]{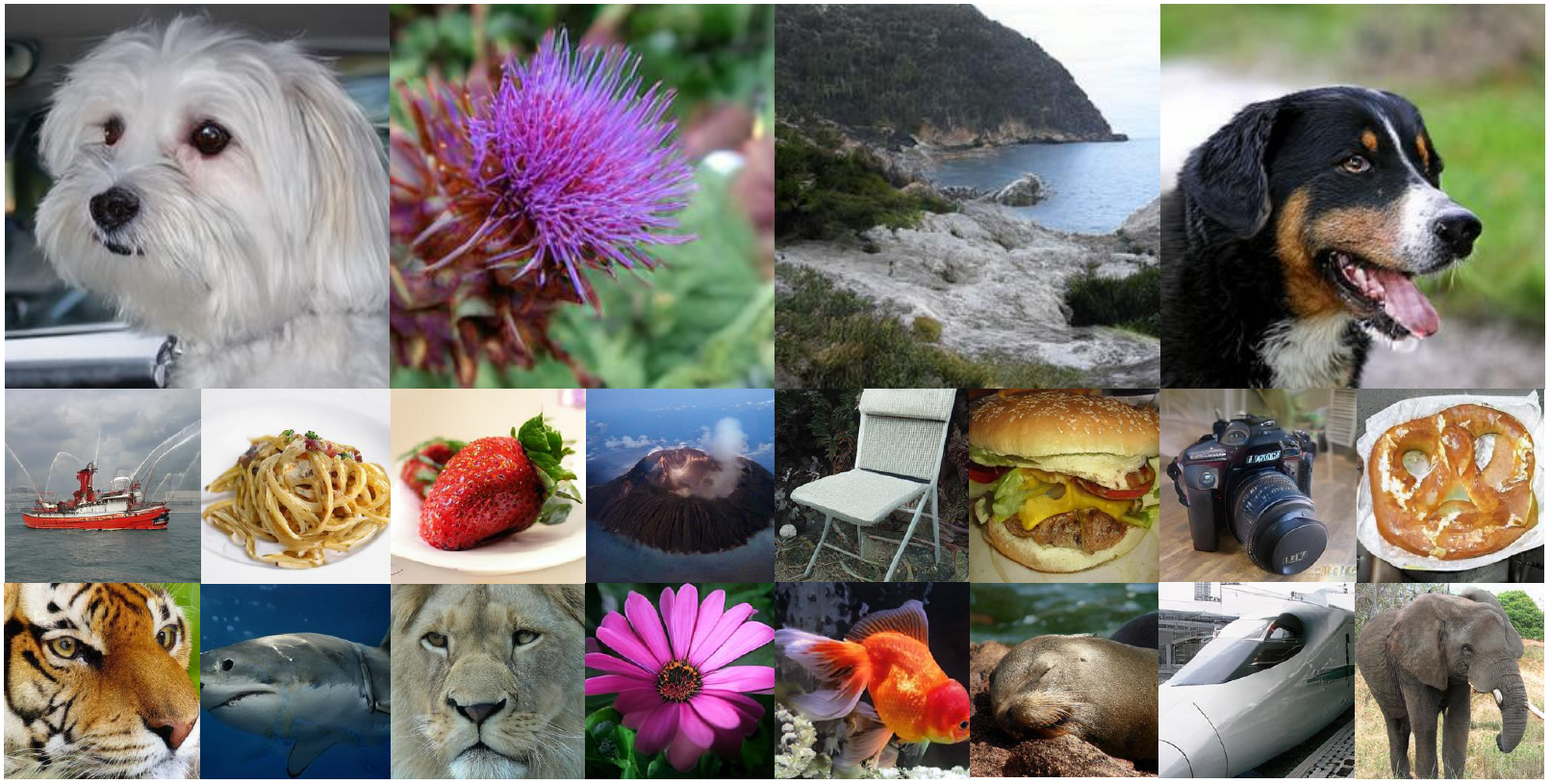}
    \vspace{-0.1in}
    \caption{ImageNet-1K 256$\times$256 generation results of generative models trained with 64 tokens. We include a versatile classes of images such as animals, plants and food. Our method achieve pleasant generation outcomes with efficient token representation and local details.
    }
    \label{fig:illustration}
    \vspace{-0.2in}
\end{figure*}
\subsection{Image tokenization}
Image tokenization methods can be broadly categorized by their latent space type (discrete vs. continuous) and spatial structure (2D grid vs. 1D sequence).
For discrete modeling, Vector Quantization (VQ) frameworks~\cite{van2017neural, esser2021taming, yu2021vector} learn a codebook of discrete visual tokens. In contrast, continuous models such as KL-VAEs~\cite{kingma2013auto} employ the reparameterization trick to constrain latents to Gaussian priors.
Most early works, including VQ-VAE and SD-VAE~\cite{rombach2022high}, adopt 2D grid latents to preserve spatial structure, a design that underpins diffusion-based models like Stable Diffusion. While highly effective, these 2D formulations impose limitations in compactness and efficiency, prompting recent research into 1D image tokenizers that represent images as highly compressed sequential embeddings.

The study of 1D image tokenization began with Titok~\cite{yu2024image}, which demonstrated effective reconstruction using only 32 tokens by representing images as short 1D sequences. Titok uses a two-stage training pipeline for discrete 1D VQ modeling, leveraging codes from pretrained models like MaskGIT-VQGAN~\cite{chang2022maskgit}. Subsequent works have explored various directions, including Flowtok and TA-Titok~\cite{he2025flowtok, kim2025democratizing} for text-to-image generation, Flextok and OneDpiece~\cite{bachmann2025flextok, miwa2025one} for variable-length sequences, and GigaTok~\cite{xiong2025gigatok} for combining 1D and 2D structures. Recent work~\cite{beyer2025highly} further investigates 1D sequences for fine-grained image editing.
To improve tokenization quality, SoftVQ~\cite{chen2025softvq} introduces a differentiable soft vector quantization mechanism, and MAE-Tok~\cite{chen2025masked} leverages masked autoencoding for semantically enriched latent spaces with strong reconstruction fidelity. However, none of these tokenizers explicitly differentiate or exploit the global-local information. To address this, we propose SelfBootTok, a 1D tokenizer that models global-local interactions for compact, efficient image representation.

\subsection{Generative Models}
Generative models are critical for evaluating image tokenizers in downstream generation tasks, and can be broadly divided into diffusion-based and autoregressive approaches.
Diffusion-based models generate images by gradually converting Gaussian noise into structured latent codes. Representative methods include DiT, SiT, and MAR~\cite{peebles2023scalable, ma2024sitexploringflowdiffusionbased, li2024autoregressive}. DiT adopts a Transformer backbone and models diffusion via stochastic differential equations. SiT leverages stochastic interpolants and optimizes the velocity field of a probability flow ODE. MAR unifies diffusion and autoregression within an encoder-decoder Transformer. Lightning-DiT~\cite{yao2025reconstruction} further boosts efficiency via a lightweight design that balances reconstruction and generation quality.
Autoregressive models follow the next-token prediction paradigm from LLMs, enabling unified multimodal generation. Methods incorporating masked modeling~\cite{chang2022maskgit, weber2024maskbit, yu2023language, yu2024image} and next-scale prediction~\cite{tian2024visual, li2024imagefolder} further improve representation and generation quality.

\subsection{Self-Bootstrapped and Scaling Paradigms in Multimodal Learning}
Self-bootstrapping has proven effective for multimodal learning by fully exploiting model capacity. For instance, BLIP-2~\cite{li2023blip} introduced it for vision--language fusion, OK-VQA~\cite{hao2024self} and VILA~\cite{fang2024vila} applied it to visual question answering and model refinement, and VideoJudge~\cite{waheed2025videojudge} extended it to scalable video assessment. Further works~\cite{xia2025bootstrapping, ding2024lowis3d} validated its potential in data-efficient reasoning and 3D scene understanding. Nevertheless, its application to image tokenization remains largely unexplored.
The success of scaling in LLMs has motivated extensive research in multimodal understanding and generation. In visual understanding, recent studies have focused on scaling vision encoders within multimodal LLMs~\cite{alayrac2022flamingo, awadalla2023openflamingo, dai2023instructblip, li2023videochat, lin2023video, maaz2023video, liu2023visual, achiam2023gpt, team2023gemini, chi2026enhancing}, with InternVideo2~\cite{wang2024internvideo2} and CuMo~\cite{li2024cumo} adopting efficient scaling and sparsely-gated Mixture-of-Experts layers. For image understanding, SViT~\cite{zhao2023svit}, LongLLaVA~\cite{wang2024longllava}, and LLaVA-scale~\cite{lu2023empirical} investigate backbone scaling, while AuroraCap~\cite{chai2024auroracap} and LLaVA-Next~\cite{li2024llava} use multi-stage training to unify visual instruction tuning across image and video.
In multimodal generation, scaling remains a critical challenge. Methods such as ViT-VQGAN~\cite{yu2021vector} and ViTok~\cite{hansenestruch2025learningsscalingvisualtokenizers} show that larger tokenizers do not always improve generation performance and often underuse model capacity. GigaTok~\cite{xiong2025gigatok} improves scaling efficiency via hybrid CNN-Transformer tokenizers. However, efficiently scaling local token information is still an open problem, which we target in this work.
\section{Method}
\begin{figure*}[th!]
    \centering
    \includegraphics[width=0.95\linewidth]{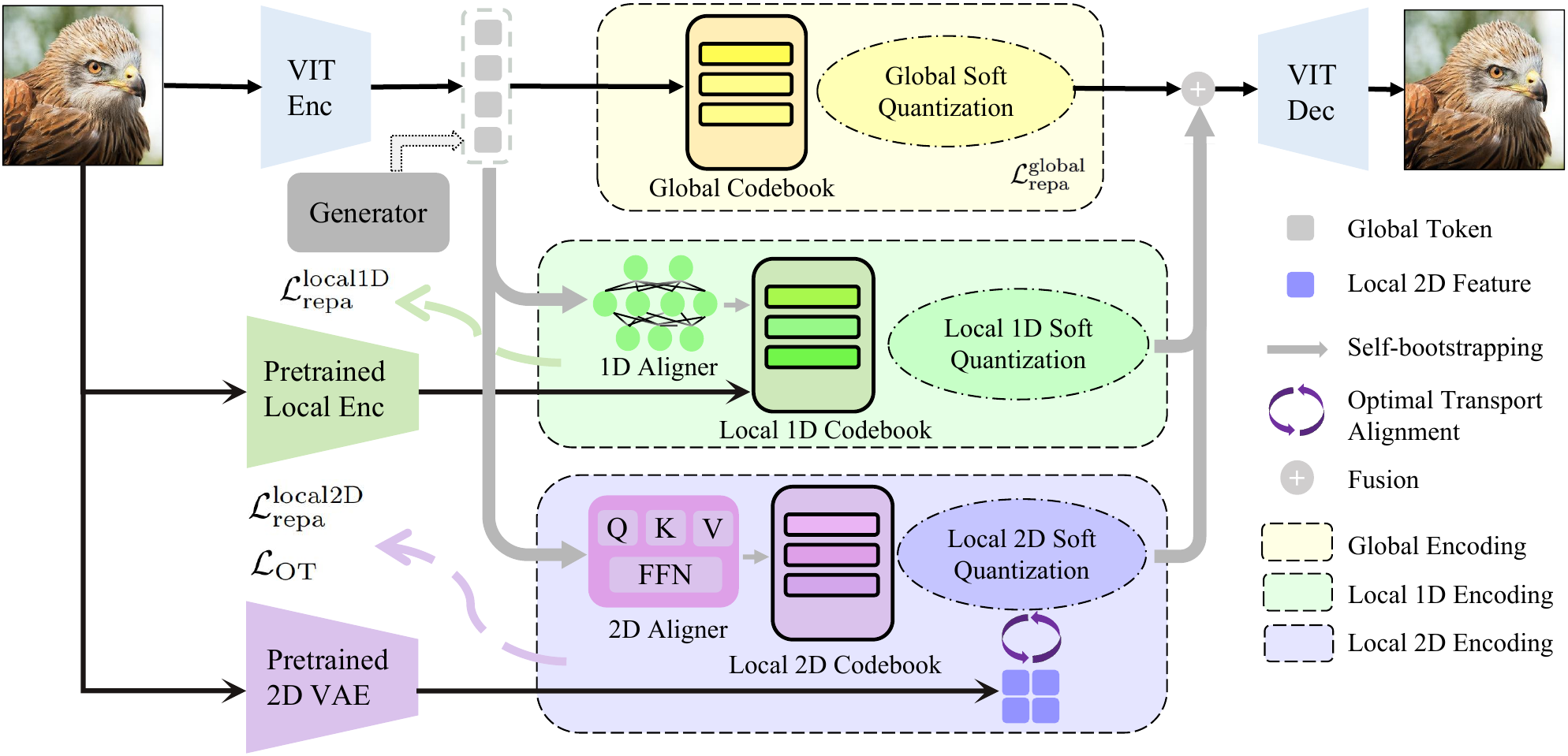}
    \vspace{-0.1in}
    \caption{Overview of the proposed SelfbootTok pipeline. The input image is first encoded into a set of global tokens using a ViT backbone. Subsequently, local tokens of varying granularity (i.e., 1D or 2D) are predicted through a self-bootstrapping paradigm. These local tokens are aligned with different pretrained visual encoders to capture multi-granularity structural information. Both global and local tokens are then softly quantized, fused, and finally decoded using a ViT decoder. This overall architecture offers strong scalability and training efficiency, enabling high-quality reconstruction and generation with minimal computational overhead.
    }
    \label{fig:teaser1}
    \vspace{-0.1in}
\end{figure*}

Our image tokenization framework consists of three key components: global-local decomposition, self-supervised bootstrapped learning, and 2D to 1D alignment via optimal transport. \textbf{Our framework employs a hierarchical self-bootstrapped VAE, which theoretically achieves superior performance (see Appendix Section A for detailed proofs).}
The global-local decomposition aims to separately optimize global and local token representations, thereby reducing redundant information and improving representational efficiency.
The self-supervised bootstrapped learning mechanism is designed to capture the intrinsic relationships between global semantics and fine-grained local details, facilitating hierarchical feature learning. The 2D to 1D alignment module preserves the sequential structure of 1D tokens by transporting the 2D token distribution into a 1D sequence space and minimizing an optimal transport loss.
Moreover, our framework naturally supports efficient scaling of local aligners, enabling flexible adaptation to different computational budgets while maintaining high generation performance and significantly reducing training cost.
\subsection{Global-local Decomposition}
As shown in~\cref{fig:teaser1}, after getting the global tokens $z_G$ from input image $I$ through a Vision Transformer encoder $Enc$, our pipeline recovers more local information through local alignment and get local 1D tokens $z_{L1}$ and local 2D tokens $z_{L2}$.
Then, we apply a multi-codebooks design that distinguishes local entries with global entries and enhances the localized representations with either 1D or 2D information injection:
\begin{equation}
\begin{aligned}
    & \textbf{Encode:} \quad 
    \mathbf{z}_G = Enc(I), \quad 
    \mathbf{z}_{\mathrm{tot}} = \{ \mathbf{z}_G, \mathbf{z}_{L1}, \mathbf{z}_{L2} \}, \\[4pt]
    & \textbf{Codebook Set:} \quad 
    \mathcal{C}_{\mathrm{tot}} = \{ 
    \mathcal{C}_{\mathrm{global}}, 
    \mathcal{C}_{\mathrm{local1D}}, 
    \mathcal{C}_{\mathrm{local2D}} 
    \}.
\end{aligned}
\label{eq:encoder_codebook}
\end{equation}

We apply soft vector quantization~\cite{chen2025softvq} such that both global and local quantizations are fully-differentiable, and thus the encoder and codebook can be optimized directly from the reconstruction loss:
{\small
\begin{align}
q_{\phi}(Z \mid I) 
&= \mathrm{Softmax}\!\left(
    -\tfrac{1}{\tau}\lVert \hat{Z} - \mathcal{C} \rVert_2
  \right),
\ \hat{Z} \!\in\! \mathbf{z}_{\mathrm{tot}}, \\[-2pt]
Z 
&= q_{\phi}(Z \mid I) \, \mathcal{C},
\ \mathcal{C} \!\in\! \mathcal{C}_{\mathrm{tot}}.
\end{align}
}

Under this design, we separate different token groups using distinct codebooks, allowing each group to specialize in learning representations of a specific granularity. The overall training objective combines reconstruction, perceptual, adversarial, alignment, and KL regularization terms:
\begin{equation}
\mathcal{L} = \mathcal{L}_{\textrm{recon}} + \lambda_1 \mathcal{L}_{\textrm{percep}} + \lambda_2 \mathcal{L}_{\textrm{adv}} + \lambda_3 \mathcal{L}_{\textrm{align}} + \lambda_4 \mathcal{L}_{\textrm{KL}}, 
\end{equation}

The alignment loss $\mathcal{L}_{\mathrm{align}}$ further decomposes into three components, corresponding to the global, local, and cross-dimensional (2D to 1D) alignment objectives:
\begin{equation}
\mathcal{L}_{\textrm{align}} =  \mu_1 \mathcal{L}_{\mathrm{repa}}^{\mathrm{global}} +  \mu_2 \mathcal{L}_{\mathrm{repa}}^{\mathrm{local}} +  \mu_3 \mathcal{L}_{\mathrm{OT}}
\label{eq:align_loss}
\end{equation}

The global alignment loss encourages global tokens to align with the features extracted by a pretrained global semantic encoder (e.g., DINOv2~\cite{oquab2023dinov2}). This loss is computed as a patch-wise similarity measure:
\begin{equation}
\mathcal{L}_{\mathrm{repa}}^{\mathrm{global}}
= \text{PatchwiseSim}(
\mathbf{z}_G, f_{\mathrm{DINOv2}}(I))
.
\label{eq:repa_global}
\end{equation}

Similarly, the local alignment loss matches each local 1D token to its corresponding localized representation from pretrained encoders (e.g., SigLIP2~\cite{tschannen2025siglip} or I-JEPA~\cite{assran2023self}):
\begin{equation}
\mathcal{L}_{\mathrm{repa}}^{\mathrm{local}} = \text{PatchwiseSim}(\mathbf{z}_{L1} \text{ or } \mathbf{z}_{L2}, f_{\mathrm{local}}(I)).
\label{eq:repa_local}
\end{equation}

Finally, $\mathcal{L}_{\mathrm{OT}}$ denotes the optimal transport loss that aligns the 2D VAE latent representations of images with their corresponding 1D token sequences, as defined in~\cref{eq:transport_loss}.
This design effectively disentangles global and local information across token groups, leading to more compact and efficient representations, particularly for global tokens.

\subsection{Self-supervised Bootstrapped Learning}
After obtaining the global tokens $\mathbf{z}_G$ from~\cref{eq:encoder_codebook}, we employ a self-bootstrapped paradigm to predict local tokens from these global representations. Specifically, to predict local 1D tokens $\mathbf{z}_{L1}$, we use an MLP module to project $\mathbf{z}_G$ into a higher-dimensional token space:

\begin{equation}
\mathbf{z}_{L1} = \mathrm{MLP}(\mathbf{z}_{G}),
\label{eq:l1_prediction}
\end{equation}

To predict local 2D tokens, we employ several scalable causal transformer blocks that autoregressively map $\mathbf{z}_G$ to another sequence of tokens $\mathbf{z}_{L2}$:
\begin{equation}
p_{\theta}(\mathbf{z}_{L2} \mid \mathbf{z}_{G})
=
\prod_{i=1}^{N_l}
p_{\theta}\big(
\mathbf{t}^{\mathrm{local}}_{i}
\mid
\mathbf{z}_{G},
\mathbf{t}^{\mathrm{local}}_{<i}
\big),
\label{eq:bootstrap_prob}
\end{equation}

We then align this token sequence with the corresponding sequence transported from the 2D VAE latents, as described in~\cref{eq:transport_loss}. During the generation process, the generator produces only the global tokens $\mathbf{z}_G$. The local tokens $\mathbf{z}_{L1}$ and $\mathbf{z}_{L2}$ are then automatically predicted through the local 1D and 2D aligners. Finally, after the soft quantization of $\mathbf{z}_G$, $\mathbf{z}_{L1}$, and $\mathbf{z}_{L2}$, all tokens are fused and decoded together:
\begin{equation}
\hat{\mathbf{x}} = \mathrm{Dec}\big(\mathrm{Fuse}(\mathbf{z}_G, \mathbf{z}_{L1}, \mathbf{z}_{L2});\mathbf{z}_{L1}; \mathbf{z}_{L2}\big),
\label{eq:generation_pipeline}
\end{equation}

The fused tokens, along with $\mathbf{z}_{L1}$ and $\mathbf{z}_{L2}$, are then injected into the cross-attention layers of the decoder. We explore two fusion strategies for decoding: soft residual fusion and concatenation fusion. Further ablation studies comparing these strategies are provided in the experimental section.

\begin{equation}
\small
\mathrm{Fuse} =
\begin{cases}
\alpha\,\mathbf{z}_G 
+ \beta\,\mathbf{z}_{L1} 
+ (1{-}\alpha{-}\beta)\,\mathbf{z}_{L2}, & \!\!\text{(Soft residual)}\\[3pt]
\mathrm{Concat}[\mathbf{z}_G;\mathbf{z}_{L1};\mathbf{z}_{L2}], & \!\!\text{(Concatenation)}
\end{cases}
\label{eq:fusion}
\end{equation}

Within our self-bootstrapped paradigm, generating only high-level, compact global information is sufficient to achieve high-quality generation. This approach reduces computational costs and optimizes the latent space of global tokens.

\subsection{2D to 1D alignment via optimal transport}
To efficiently inject 2D priors into a 1D token sequence and achieve a compact 1D representation, we use a local 2D aligner to predict a 1D sequence and align the 2D VAE latent features with it. \textbf{This formulation naturally fits into the optimal transport (OT) framework, which constructs mappings between tensors of different dimensions while encouraging compact representations, and can be efficiently solved via the Sinkhorn algorithm.} Specifically, let \(\mathbf{X}\in\mathbb{R}^{B\times N_x\times d}\) be the matrix of 2D VAE latents
(with \(N_x=H\! \times\! W\) spatial tokens),
and \(\mathbf{z}_{L2}\in\mathbb{R}^{B \times N_z\times d}\) be the matrix of 1D tokens
(with \(N_z\) latent tokens). Here, \(B\) denotes the batch size, and \(d\) refers to the token dimension. The goal is to find a transport plan
\(\mathbf{P}\in\mathbb{R}_{+}^{N_x\times N_z}\) that couples the marginals
\(\mathbf{a}\in\Delta^{N_x}\) and \(\mathbf{b}\in\Delta^{N_z}\) (where
\(\Delta^N\) denotes the \(N\)-simplex).

Let the squared-Euclidean cost matrix be
\[
\mathbf{C}_{ij} = \lVert \mathbf{x}_i - \mathbf{z}_j \rVert_2^2, \quad \text{for } i \in [1, N_x], \ j \in [1, N_z]
\]
measuring pairwise distances between 2D latents \(\mathbf{X}\) and 1D tokens \(\mathbf{z}_{L2}\). We solve
the entropy-regularized optimal transport problem
\begin{equation}
\mathbf{P}^\ast
= \arg\min_{\mathbf{P}\in\Pi(\mathbf{a},\mathbf{b})}
\ \langle \mathbf{P}, \mathbf{C}\rangle \;+\; \varepsilon\,\mathcal{H}(\mathbf{P}),
\label{eq:ot_objective}
\end{equation}
where \(\Pi(\mathbf{a}, \mathbf{b})\) is the set of transport plans coupling \(\mathbf{a}\) and \(\mathbf{b}\), \(\langle \mathbf{.}, \mathbf{.} \rangle\) represents the Frobenius inner product and \(\mathcal{H}(\mathbf{P})\) is the entropy regularizer. The optimal transport plan \(\mathbf{P}^\ast\) is computed using the Sinkhorn algorithm:
\[
\mathbf{P}^\ast = \mathrm{Sinkhorn}(\mathbf{K}, \mathbf{a}, \mathbf{b}),
\]
where \(\mathbf{K} = \exp(-\mathbf{C}/\varepsilon)\). 
And we use the transport cost as part of the alignment loss shown in~\cref{eq:align_loss}:
\begin{equation}
\mathcal{L}_{\mathrm{OT}} \;=\; \langle \mathbf{P}^\ast, \mathbf{C}\rangle.
\label{eq:transport_loss}
\end{equation}

\begin{figure}[t!]
    \centering
    \includegraphics[width=0.7\linewidth]{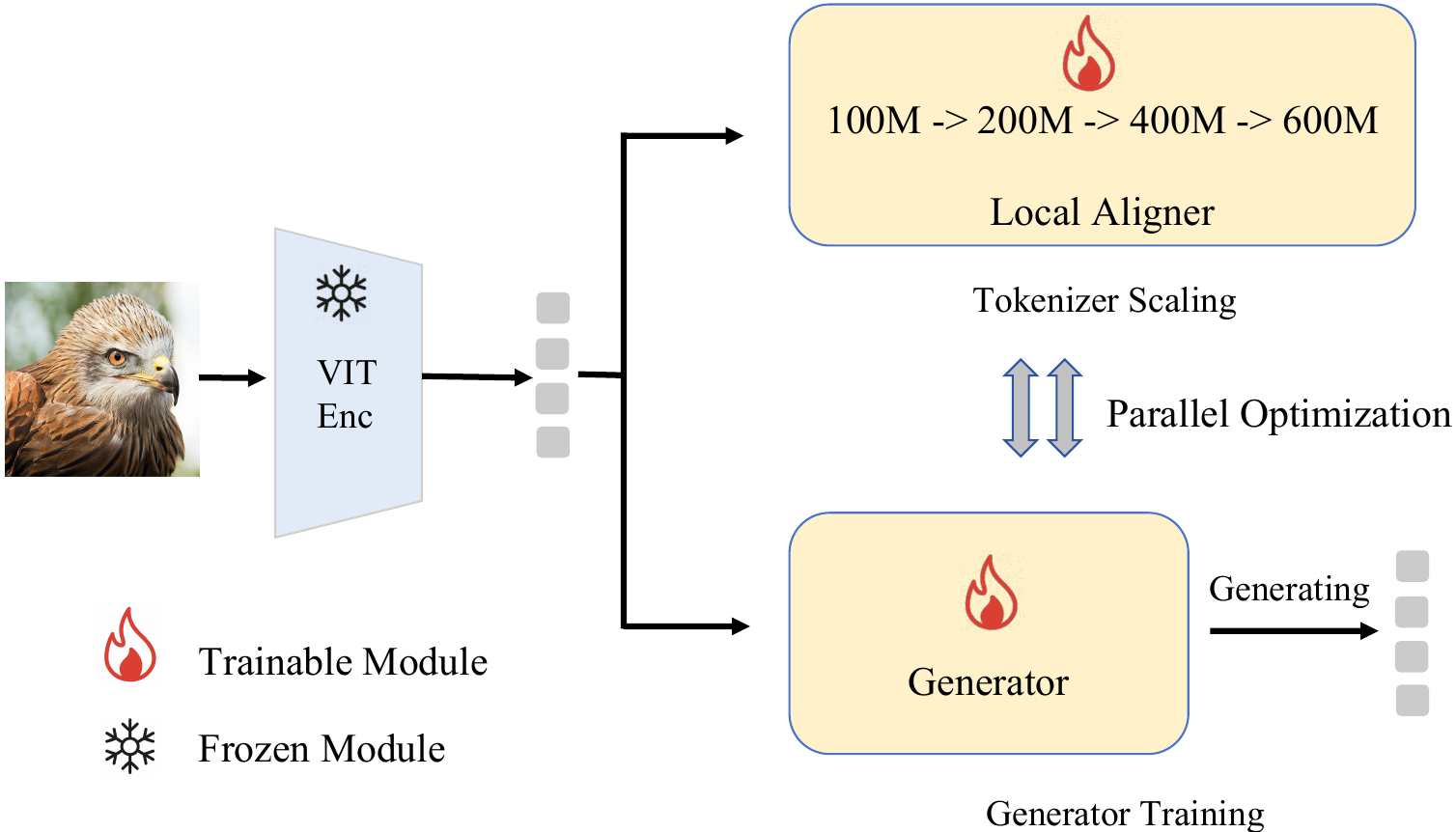}
    \vspace{-0.1in}
    \caption{Illustration of parallel training pipeline. Our method supports training larger scale tokenizer and generator at the same time, which enhances training efficiency.}
    \label{fig:architecture22}
    \vspace{-0.1in}
\end{figure}
\begin{table*}[t]
\centering
\renewcommand{\arraystretch}{1.15}  
\setlength{\tabcolsep}{5.5pt}       
\scriptsize                         
\begin{tabular}{lccccc}
\toprule
\textbf{Tokenizer} & \textbf{\#Tokens} & \textbf{Generators} & \textbf{rFID$\downarrow$} & \textbf{gFID(w/o CFG)$\downarrow$} & \textbf{gFID(w CFG)$\downarrow$} \\
\midrule[\heavyrulewidth]  
\multicolumn{6}{l}{\textit{\textbf{2D Tokenizer}}} \\  
\midrule
VAE~\cite{rombach2022high} & 4096 & LDM-4~\cite{rombach2022high}  & 0.27 & 10.56 & 3.60 \\
SD-VAE~\cite{rombach2022high} & 256 & DiT-XL/2~\cite{peebles2023scalable}  & 0.62 & 9.62 & 2.27 \\
SD-VAE~\cite{rombach2022high} & 256 & SiT-XL/2~\cite{ma2024sitexploringflowdiffusionbased}  & 0.62 & 8.61 & 2.06 \\
VAR-Tok~\cite{tian2024visual} & 680 & VAR-d30~\cite{tian2024visual} & 1.00 & -- & 1.92 \\
VQGAN~\cite{esser2021taming} & 256 & LlamaGen~\cite{sun2024autoregressive} & 0.59 & 9.38 & 2.18 \\
VA-VAE~\cite{yao2025reconstruction} & 256 & LightningDiT~\cite{yao2025reconstruction}  & 0.28 & 2.17 & 1.35 \\
\midrule[\heavyrulewidth]  
\multicolumn{6}{l}{\textit{\textbf{Auto-regressive Tokenizer}}} \\
\midrule
MaskGIT~\cite{chang2022maskgit} & 256 & MaskGIT~\cite{chang2022maskgit}  & 2.28 & 6.18 & -- \\
LlamaGen-3B~\cite{sun2024autoregressive} & 576 & LlamaGen~\cite{sun2024autoregressive} & 2.19 & -- & 2.18 \\
RQ-VAE~\cite{lee2022autoregressive} & 256 & RQ-Transformer~\cite{lee2022autoregressive} & 3.20 & -- & 3.80 \\
MAGE~\cite{li2023mage} & 256 & MAGE~\cite{li2023mage} & -- & 6.93 & -- \\
MAR-H~\cite{li2024autoregressive} & 256 & MAR-H~\cite{li2024autoregressive} & 1.22 & 2.35 & 1.55 \\
ViTok~S-B/16~\cite{hansenestruch2025learningsscalingvisualtokenizers} & 256 & DiT-XL~\cite{peebles2023scalable}  & -- & -- & 2.45 \\
\midrule[\heavyrulewidth]  
\multicolumn{6}{l}{\textit{\textbf{1D Tokenizer}}} \\
\midrule
TiTok-B~\cite{yu2024image} & 64 & MaskGIT-ViT~\cite{chang2022maskgit}  & 1.70 & -- & 2.48 \\
TiTok-S~\cite{yu2024image} & 128 & MaskGIT-UViT-L ~\cite{chang2022maskgit, bao2023all}  & 1.71 & -- & 1.97 \\
FlexTok~d18-d28~\cite{bachmann2025flextok} & 1 to 256 & LlamaGen~\cite{sun2024autoregressive}  & 1.45 & -- & 1.86 \\
OneD-piece~\cite{miwa2025one} & 1 to 256 & -- & 1.08 & -- & -- \\
HieraTok~\cite{chen2025hieratok} & 256 & DiT~\cite{peebles2023scalable}  & 0.45 & 3.53 & 1.82 \\
GigaTok~XL-XXL~\cite{xiong2025gigatok} & 256 & LlamaGen~XXL~\cite{sun2024autoregressive} & 0.79 & -- & 1.98 \\
MAETok~\cite{chen2025masked} & 128 & LightningDiT~\cite{yao2025reconstruction} & 0.48 & -- & 1.73 \\
SoftVQ~\cite{chen2025softvq} & 64 & SiT~\cite{ma2024sitexploringflowdiffusionbased} & 0.88 & 5.98 & 1.78 \\
SelfBootTok & \textbf{64} & SiT~\cite{ma2024sitexploringflowdiffusionbased}  & 0.66 & 2.22 & 1.56 \\
\bottomrule
\end{tabular}
\caption{\textbf{Performance of SelfBootTok} in the context of state-of-the-art generative models on ImageNet-256. 
We additionally report the number of latent tokens (\#Tokens) for each tokenizer. Our method achieves state-of-the-art generation performance gfid 1.56 among 1D tokenizers with only 64 tokens.}
\label{tab:scaleup-results}
\end{table*}
\subsection{Local Scaling with heavier local aligner}
\begin{figure}[t!]
    \centering
    \begin{minipage}{0.48\textwidth}
        \centering
        \includegraphics[width=\linewidth]{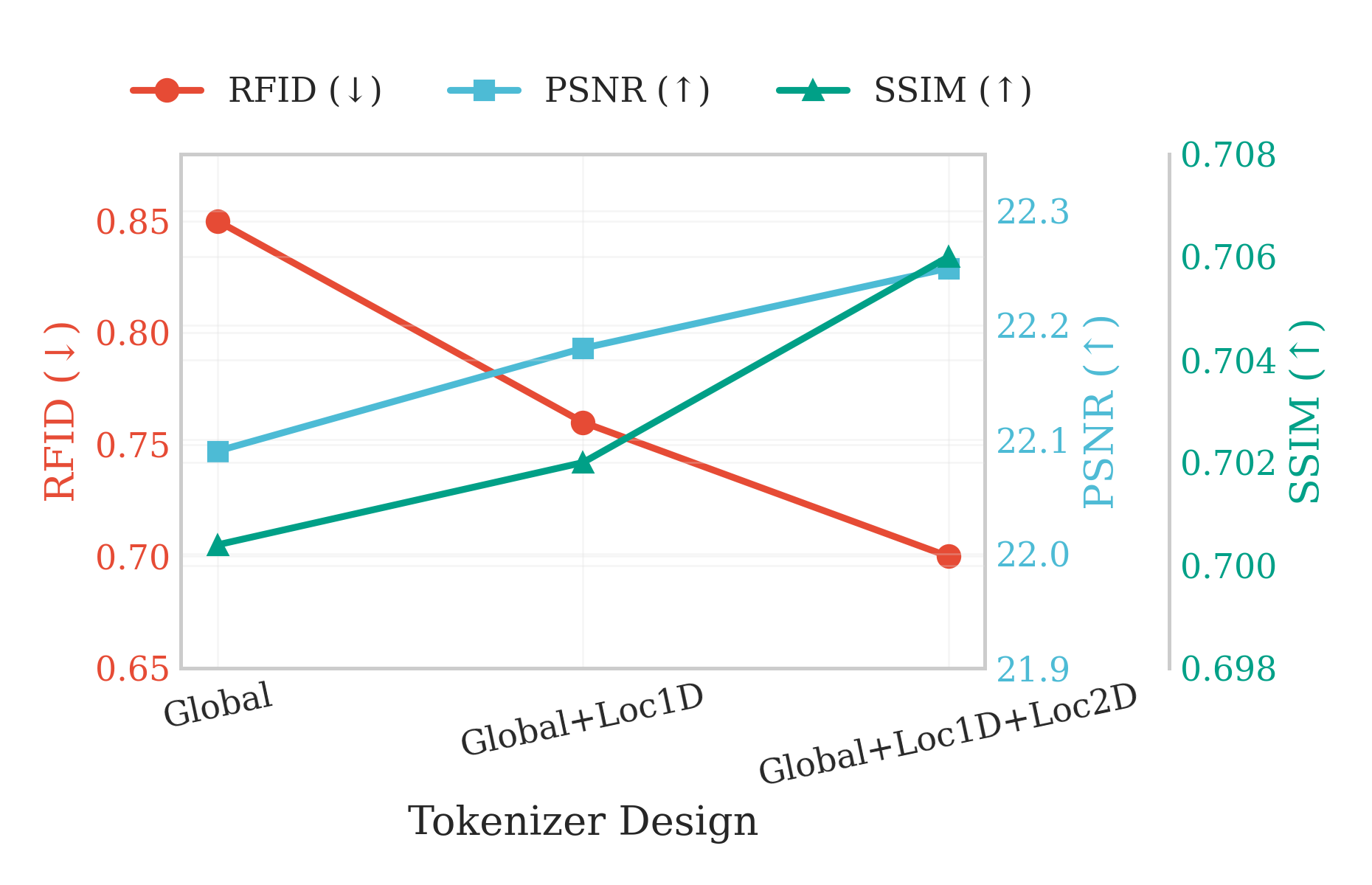}
        \caption{Ablation study of global-local self-bootstrapping design. RFID, PSNR, and SSIM curves are reported. As shown, incorporating self-bootstrapped local 1D and 2D tokens substantially improves reconstruction performance across all metrics.}
        \label{fig:architecture}
    \end{minipage}
    \hfill  
    \begin{minipage}{0.48\textwidth}
        \centering
        \includegraphics[width=\linewidth]{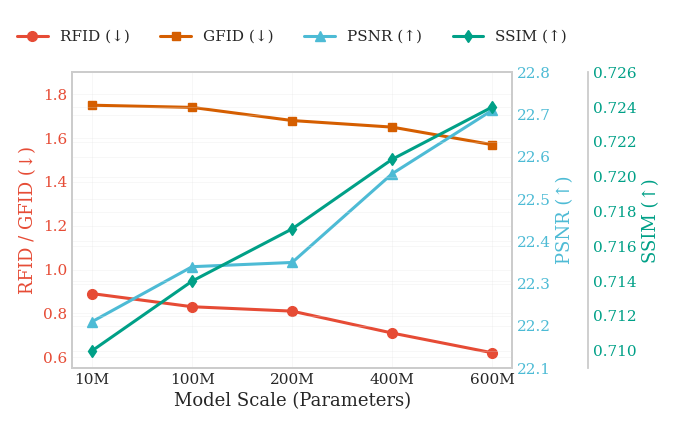}
        \caption{Scaling results for the local 2D aligner are presented, where the model size is scaled from 10M to 600M. As shown, efficient scaling improves both reconstruction and generation quality. Notably, for the five generation results (gFID) shown, the generator is trained only once, resulting in a 40 percent reduction in total computational cost.}
        \label{fig:architecture11}
    \end{minipage}
    \vspace{-0.1in}
\end{figure}

To explore the potential of the self-bootstrapped paradigm, we conduct experiments using local aligners at different spatial scales. Specifically, we vary the size of the local 2D module while keeping all other components fixed. When the local 2D aligner becomes large, the overall model becomes computationally heavy and difficult to optimize. To address this issue, we adopt a two-stage training strategy. In the first stage, we train the entire pipeline with a smaller local 2D aligner. In the second stage, we scale up the local 2D aligner and fine-tune it while freezing the remaining components. This strategy allows simultaneous training of the generator and the enlarged local aligner, enabling the generator to be trained once for global tokens and subsequently reused across different local aligner scales. Further discussions on the scaling results are provided in Section~4.2 and~\cref{fig:architecture11}.

\section{Experiments}
\begin{table*}[t]
\centering
\begin{minipage}[t]{0.42\textwidth}
    \centering
    \setlength{\tabcolsep}{5pt}
    \renewcommand{\arraystretch}{1.1}
    \caption{\textbf{Ablation study of different local1D aligners.} We compare reconstruction performance using I-JEPA, DINOv2, and SigLIP as 1D alignment encoders, evaluated by rFID, PSNR, and SSIM metrics under the 400M local 2D aligner.}
    \vspace{3pt}
    \begin{small}
    \begin{tabular}{lccc}
    \toprule
    \textbf{Aligner} & \textbf{rFID ($\downarrow$)} & \textbf{PSNR ($\uparrow$)} & \textbf{SSIM ($\uparrow$)} \\
    \midrule
    I-JEPA & \textbf{0.66} & \textbf{22.86} & \textbf{0.7288} \\
    DINOv2 & 0.70 & 22.68 & 0.7192 \\
    SigLIP & 0.72 & 22.45 & 0.7150 \\
    \bottomrule
    \end{tabular}
    \end{small}
    \vspace{-3pt}
    \label{tab:local1d_ablation}
\end{minipage}
\hfill
\begin{minipage}[t]{0.54\textwidth}
    \centering
    \setlength{\tabcolsep}{6pt}
    \setlength{\abovecaptionskip}{2pt}
    \setlength{\belowcaptionskip}{0pt}
    \renewcommand{\arraystretch}{1.35}
    \caption{\textbf{Ablation results of different design choices} (600M parameter tokenizer, 64 tokens). G: global; L1/L2: local 1D/2D; C: causal; CB: local codebooks.}
    \begin{tabular}{lcccc}
    \toprule
    \textbf{Method} & \textbf{C} & \textbf{CB} & \textbf{gFID} & \textbf{rFID} \\
    \midrule
    SoftVQ        & -- & -- & 1.78 & 0.88 \\
    G + L1        & --  & $\checkmark$ & 1.70 & 0.76 \\
    G + L1 + L2   & $\times$  & $\checkmark$  & 1.65  & 0.72   \\
    G + L1 + L2   & $\checkmark$  & $\times$  & 1.68  & 0.73   \\
    G + L1 + L2   & $\checkmark$  & $\checkmark$  & 1.56  & 0.66 \\
    \bottomrule
    \end{tabular}
    \vspace{-4pt}
    \label{tab:ablation1111}
\end{minipage}
\end{table*}

\subsection{Experiments Setup}
\noindent \textbf{Implementation Details of SelfBootTok}. 
In this section, we introduce the experiment settings of our method. Our tokenizer utilizes Dinov2 VIT~\cite{oquab2023dinov2} as the backbone and the encoder is initialized the same as pretrained Dinov2 encoder. All the experiments are conducted on the base size of Dinov2 VIT backbone with a total of 173M parameters, while the sizes of local aligners vary during the scaling experiments, including sizes of 10M, 100M, 200M, 400M and 600M. Most of our experiments are conducted with the number of latent code N = 64 and token dimension 32, while we also include comparisons of N = 32 for ablation study (See Appendix for more details). The tokenizer is trained on ImageNet at a resolution of 256$\times$256 for 50 epochs with a small local aligner, and up to 75 epochs with larger local aligners in the scaling experiments. For discriminator training, we adopt a setup similar to SoftVQ~\cite{chen2025softvq}, using a StyleGAN-like architecture~\cite{karras2019style, karras2020analyzing} and strategies such as LeCAM~\cite{tseng2021regularizing} and consistency regularization~\cite{zhang2019consistency}.

\noindent \textbf{Implementation Details of Generators}.
We use SiT as the generator for downstream denoising-based image generation tasks. In our setup, where only global tokens need to be generated, we train our tokenizer for 50 epochs and SiT for 800K iterations. For subsequent scaling, we expand the local 2D aligner and train the tokenizer pipeline for an additional 50 to 75 epochs, keeping the ViT encoder frozen. Notably, SiT requires no further training. To accelerate SiT's training, we apply the representation disentangling method~\cite{wu2025representation} and incorporate a pretrained DINOv2-base encoder~\cite{oquab2023dinov2} to enhance convergence speed.

\noindent \textbf{Evaluation}.
We adopt reconstruction Frechet Inception Distance (rFID)~\cite{heusel2017gans} on ImageNet validation set to assess the performance of tokenizer. Furthermore, we include PSNR and SSIM metrics to more accurately reflect the reconstruction quality. To evaluate the generation quality, we report generation FID (gFID) under either with or without CFG setting. Additionally, we evaluate the efficiency of generative models with respect to tokenizer GFLOPs. For further details, refer to the Appendix.

\subsection{Quantitative Results}
We analyze the reconstruction and generation results, along with the training efficiency in this section.

\noindent \textbf{High reconstruction quality with only 64 tokens.}
As shown in~\cref{tab:scaleup-results}, our method achieves superior reconstruction performance (\ie, an rFID score of \textbf{0.66}) using only 64 tokens, outperforming other baselines trained with the same token budget~\cite{chen2025softvq, yu2024image} and performing comparably to methods~\cite{esser2021taming, rombach2022high, xiong2025gigatok} that rely on 256 tokens. Moreover, after efficient scaling, our method achieves better PSNR and SSIM scores for reconstruction, as illustrated in~\cref{fig:architecture11}. This demonstrates the potential for progressively improved reconstruction performance as the local aligner is scaled up within our architecture.

\noindent \textbf{State-of-the-art generation performance among 1D tokenizers.}
In terms of generation, our approach achieves a state-of-the-art gFID score of \textbf{1.56} using only 64 tokens, as reported in~\cref{tab:scaleup-results}. This result is comparable to the best-performing 2D tokenizer VA-VAE (\ie, 1.35 gFID with 256 tokens) and the top autoregressive tokenizer MAR-H (\ie, 1.55 gFID with 256 tokens). Moreover, our model attains a lower gFID even without classifier-free guidance (CFG), outperforming other 1D and autoregressive baselines under the same setting. This shows that incorporating multi-granularity information into our model architecture enhances generation quality, leading to superior performance and validating the effectiveness of our design.

\noindent \textbf{Efficient training.}
Our method also exhibits strong training efficiency, achieving competitive performance without additional generation optimization overhead. As illustrated in~\cref{fig:architecture22}, once the tokenizer pipeline is trained, our framework supports parallel optimization of the scaled local aligners and the generator. Unlike conventional approaches that require retraining the generator each time the tokenizer is scaled, our design leverages global tokens for generation, allowing the generator to be trained only once. For the overall scaling experiment, this strategy reduces training computation by about \textbf{40\%} (from 4.9B parameters and 60K GFLOPs to 2.95B and 36.6K GFLOPs) and shortens training time by roughly \textbf{54\%} (from 28 to 13 days). Moreover, our method scales efficiently to larger datasets, yielding greater computational savings as the generator size increases. Further details on computational analysis and data scaling are provided in the Appendix.

\subsection{Qualitative Results}
As shown in~\cref{fig:illustration}, our method generates diverse, high-quality real-world images using only 64 tokens, covering animals (e.g., dogs, tigers), food (e.g., noodles, burgers), and plants (e.g., flowers) sampled from 1000 ImageNet~\cite{deng2009imagenet} classes. Additionally,~\cref{fig:reconstruction1122} demonstrates that efficient scaling of the local 2D aligner enhances reconstruction performance, recovering fine local details (e.g., human facial features, animal fur texture/color).

\subsection{Ablation Study}
To justify our design choices and validate the self-bootstrapping paradigm, we conduct comprehensive ablation studies on local 1D/2D branches, 1D aligners, and fusion strategies. As shown in~\cref{tab:ablation1111}, our design outperforms all variants: we use an MLP for the efficient 1D local aligner (not scaled, captures coarse local info) and a transformer-based 2D local aligner (scaled, models rich spatial details). ~\cref{fig:architecture} illustrates that progressive integration of local structural info enhances reconstruction quality, with optimal performance when both local branches are included.
We visualize token correspondence in~\cref{fig:oneonecol} to justify our global-local decomposition: global tokens encode overall semantics, while local 1D/2D tokens capture object configurations, background structure, and fine-grained silhouettes.
~\cref{tab:local1d_ablation} investigates local 1D token alignment with frozen visual encoders (I-JEPA, DINOv2, SigLIP): I-JEPA performs best, with DINOv2 and SigLIP comparable, indicating distinct local structural cues affect tokenizer behavior.
In~\cref{tab:fusion_ablation1122}, we compare fusion strategies under Eq.~\ref{eq:fusion}: concatenation-based fusion improves reconstruction but increases decoder size/computational cost, degrading generation quality and revealing a task-dependent trade-off.
Finally,~\cref{fig:architecture11} shows our method improves with local aligner scaling: at 600M parameters and 64 tokens, it achieves a gFID of 1.56, outperforming baselines and demonstrating architectural scalability.
\section{Conclusion}
In this paper, we propose SelfBootTok, an efficient 1D image tokenizer that decomposes image tokens into global and local groups and leverages a self-supervised bootstrapped paradigm to recover sufficient local information from global tokens. Our architecture reduces image token overlap, achieving higher reconstruction quality and better efficiency in downstream generation tasks, while its local self-bootstrapped prediction design enables strong scalability and highlights self-bootstrapping's potential for image tokenization.

\bibliographystyle{plain}
\bibliography{bibs/main}

\newpage
\appendix

\section{Theoretical Analysis of Hierarchical Self-Bootstrapped VAE.}
\label{app:theory}

\subsection{Preliminaries}

We briefly recall the standard VAE objective and its hierarchical extension.  
Let $x \in \mathcal{X}$ denote an observed image, and let the latent variables
be $z_G \in \mathcal{Z}_G$ (global tokens) and $z_L \in \mathcal{Z}_L$ (local tokens).  
A hierarchical generative model factorizes as
\begin{equation}
p_\theta(x, z_G, z_L) = p_\theta(x \mid z_G, z_L)\, p_\theta(z_L \mid z_G)\, p_\theta(z_G).
\end{equation}
The variational posterior is factorized as
\begin{equation}
q_\phi(z_G, z_L \mid x) = q_\phi(z_G \mid x)\, q_\phi(z_L \mid x),
\end{equation}
which may still allow cross-level dependency through shared encoders or alignment terms.

The hierarchical VAE optimizes the evidence lower bound (ELBO)
\begin{align}
\label{eq:hierarchical_elbo}
\mathcal{L}_B(x)
&= \E_{q_\phi(z_G, z_L \mid x)}
   [\log p_\theta(x \mid z_G, z_L)]  \nonumber\\
&\quad - \KL(q_\phi(z_G \mid x)\|p_\theta(z_G))  \nonumber\\
&\quad - \E_{q_\phi(z_G \mid x)}
   \KL(q_\phi(z_L \mid x)\|p_\theta(z_L \mid z_G)).
\end{align}

The baseline single-latent VAE assumes a flat structure with latent variable $z \in \mathcal{Z}$,
\begin{equation}
\label{eq:flat_elbo}
\mathcal{L}_A(x)
= \mathbb{E}_{q_\psi(z \mid x)}[\log p_\psi(x \mid z)]
- \mathrm{KL}(q_\psi(z \mid x) \,\|\, p_\psi(z)).
\end{equation}

\subsection{Generative Assumptions}

\begin{assumption}[True generative process]
\label{asm:true_process}
The true data distribution $p^\star(x)$ factorizes through latent variables
$z_G^\star, z_L^\star$ as
\begin{equation}
p^\star(x) = \int p^\star(x \mid z_G, z_L)\, p^\star(z_L \mid z_G)\, p^\star(z_G)\, dz_L\, dz_G.
\end{equation}
\end{assumption}
\begin{assumption}[Approximation quality]
\label{asm:approx_quality}
The conditional prior in the hierarchical model $p_\theta(z_L \mid z_G)$ approximates
the true conditional $p^\star(z_L \mid z_G)$ with an expected KL error
\begin{equation}
\delta = \mathbb{E}_{p^\star(z_G)}\big[
    \mathrm{KL}\big(p^\star(z_L \mid z_G)\,\|\,p_\theta(z_L \mid z_G)\big)
\big].
\end{equation}
\end{assumption}
\begin{assumption}[Decoder regularity]
\label{asm:lipschitz}
The decoder $p_\theta(x \mid z_G, z_L)$ is $L$-Lipschitz in $(z_G, z_L)$ and has bounded log-likelihood:
$|\log p_\theta(x \mid z_G, z_L)| \le M$ for all $(x,z_G,z_L)$.
\end{assumption}

\subsection{Main Proposition}

\begin{proposition}[ELBO Advantage of Hierarchical Decomposition]
\label{prop:elbo_bound}
Under Assumptions~\ref{asm:true_process},
there exists a constant $C = O(LM)$ such that
\begin{equation}
\sup_{\phi,\theta} \mathcal{L}_B(x)
\;\ge\;
\sup_{\psi} \mathcal{L}_A(x)
- C \,\delta,
\end{equation}
where $\mathcal{L}_A$ and $\mathcal{L}_B$ are defined in
Eqs.~\eqref{eq:flat_elbo} and \eqref{eq:hierarchical_elbo}, respectively.
In other words, when the hierarchical conditional prior
$p_\theta(z_L \mid z_G)$ closely matches the true dependency between global and local factors
($\delta \approx 0$), the hierarchical model achieves an equal or tighter ELBO
than the monolithic VAE.
\end{proposition}

\subsection{Proof Sketch}

\paragraph{Step 1: Idealized hierarchical ELBO.}
Suppose we can set $p_\theta = p^\star$ and $q_\phi = p^\star(\cdot\mid x)$, then
\begin{align}
\mathcal{L}_B^\star(x)
&= \log p^\star(x)
 - \E_{p^\star(z_G|x)}
   \KL\!\big(p^\star(z_L|x,z_G)\,\|\,p^\star(z_L|z_G)\big) \nonumber\\
&\ge \log p^\star(x).
\end{align}
since the KL term is non-negative and the ELBO is tight at equality.

\paragraph{Step 2: Decomposition of modeling error.}
For any parameterization $(\phi,\theta)$,
\begin{align}
\log p^\star(x) - \mathcal{L}_B(x)
&= \KL\!\big(q_\phi(z_G,z_L\!\mid\!x)\,\|\,p_\theta(z_G,z_L\!\mid\!x)\big) \nonumber\\
&\le \KL(q_G\|p_\theta(z_G))
   + \E_{q_G}\KL(q_L\|p_\theta(z_L\!\mid\!z_G)).
\end{align}

Substituting the approximation quality of the conditional prior (Assumption~\ref{asm:approx_quality})
gives an additive deviation bounded by $\delta$ up to a constant scaling $C$ from the decoder's smoothness.

\paragraph{Step 3: Comparison with single-latent model.}
The single-latent VAE must represent both global and local variations in a single
latent vector $z$. Its optimal ELBO can be expressed as
\[
\sup_{q_\psi,\theta_\psi}\mathcal{L}_A(x)
= \log p^\star(x) - \inf_{q_\psi,\theta_\psi} \mathrm{KL}\big(
q_\psi(z\mid x)\,\|\,p_\psi(z\mid x)
\big).
\]
Since any $p_\psi(z)$ can be viewed as marginalizing $p_\theta(z_G,z_L)$ under a coupling $z = f(z_G,z_L)$,
but with information loss (by the data-processing inequality),
the marginal model $p_\psi$ cannot achieve a smaller KL to the true posterior
than the factorized one. Therefore,
\[
\sup \mathcal{L}_B(x)
\ge \sup \mathcal{L}_A(x) - C\delta.
\]
$\square$

\subsection{Discussion and limitation}
Although our method improves tokenization quality and introduces an efficient scaling paradigm for image generation, the fusion of global and local tokens demands careful architectural design, especially as token counts increase. Further work is needed to address this limitation.

\subsection{Discussion of proof}

The proposition above provide a theoretical justification for the proposed
global--local, self-bootstrapped tokenizer. When the true generative process admits a hierarchical factorization, explicitly modeling the conditional prior yields a tighter or equal ELBO bound up to the conditional modeling error $\delta$.

\section{Detailed Optimal Transport Formulation for 2D to 1D Alignment}
\label{app:ot_alignment}

This section provides a complete theoretical formulation of the 2D--to--1D
alignment process using entropy-regularized optimal transport (OT), the existence
and uniqueness of the Sinkhorn solution, and the differentiability properties
required for end-to-end training.

\subsection{Problem Setup}

Let $\mathbf{X}\in\mathbb{R}^{N_x\times d}$ denote the 2D VAE latent tokens
($N_x = H\times W$), and let $\mathbf{Z}\in\mathbb{R}^{N_z\times d}$ denote
the predicted 1D latent tokens.  
We consider probability vectors
\[
\mathbf{a}\in\Delta^{N_x},
\qquad
\mathbf{b}\in\Delta^{N_z},
\]
where $\Delta^n=\{\mathbf{u}\in\mathbb{R}_+^n: \sum_i u_i = 1\}$.

We define the quadratic ground cost
\begin{equation}
\label{eq:cost_matrix}
C_{ij} = \lVert \mathbf{x}_i - \mathbf{z}_j\rVert_2^2,
\qquad
1\le i\le N_x,\; 1\le j\le N_z.
\end{equation}

Let $\Pi(\mathbf{a},\mathbf{b})$ denote the transport polytope:
\[
\Pi(\mathbf{a},\mathbf{b})
:= \bigl\{
\mathbf{P}\in\mathbb{R}_+^{N_x\times N_z}:\ 
\mathbf{P}\mathbf{1}=\mathbf{a},\
\mathbf{P}^\top\mathbf{1}=\mathbf{b}
\bigr\}.
\]

\subsection{Entropy-Regularized Optimal Transport}

The alignment objective is formulated as the entropy-regularized OT problem
\begin{equation}
\label{eq:ot_objective_appendix}
\mathbf{P}^\star
=
\arg\min_{\mathbf{P}\in\Pi(\mathbf{a},\mathbf{b})}
\Bigl[
    \langle \mathbf{P}, \mathbf{C}\rangle
    + \varepsilon\,\mathcal{H}(\mathbf{P})
\Bigr],
\end{equation}
where the entropy term is
\[
\mathcal{H}(\mathbf{P}) = \sum_{i,j} P_{ij}(\log P_{ij}-1).
\]

The regularization parameter $\varepsilon>0$ ensures strict convexity, which guarantees a unique minimizer.

\paragraph{Fenchel Dual Form.}
The dual of~\eqref{eq:ot_objective_appendix} is
\[
\max_{\alpha,\beta\in\mathbb{R}}
\ \alpha^\top\mathbf{a} + \beta^\top\mathbf{b}
    - \varepsilon\sum_{i,j}
        \exp\!\left(
        \frac{\alpha_i+\beta_j-C_{ij}}{\varepsilon}
        \right).
\]
Strong duality holds because the primal is strictly convex over a compact polytope.

\subsection{Sinkhorn Solution}

Define the Gibbs kernel
\begin{equation}
\label{eq:kernel_def}
K_{ij} = \exp(-C_{ij}/\varepsilon).
\end{equation}

The Sinkhorn algorithm alternates row and column normalizations:
\[
\mathbf{P}^{(t)}
= 
\mathrm{diag}(\mathbf{u}^{(t)})\,\mathbf{K}\,
\mathrm{diag}(\mathbf{v}^{(t)}),
\]
with updates
\[
\mathbf{u}^{(t+1)} = \frac{\mathbf{a}}{\mathbf{K}\mathbf{v}^{(t)}},
\quad
\mathbf{v}^{(t+1)} = \frac{\mathbf{b}}{\mathbf{K}^\top\mathbf{u}^{(t+1)}}.
\]

\paragraph{Theorem 1 (Existence and Uniqueness).}
\emph{For any strictly positive kernel $\mathbf{K}$ and marginals
$\mathbf{a},\mathbf{b}\in\Delta$, the Sinkhorn iterations converge to unique scaling vectors $(\mathbf{u}^\star,\mathbf{v}^\star)$ such that
\[
\mathbf{P}^\star = \mathrm{diag}(\mathbf{u}^\star)\mathbf{K}\mathrm{diag}(\mathbf{v}^\star)
\]
is the unique minimizer of the entropy-regularized OT problem~\eqref{eq:ot_objective_appendix}.}

\emph{Proof Sketch.}
The kernel $\mathbf{K}$ is strictly positive by~\eqref{eq:kernel_def}.
Entropy regularization makes the objective strictly convex on the transport polytope.
By the Sinkhorn--Knopp theorem, the iterative proportional fitting procedure converges to
the unique matrix in the matrix scaling set matching $\mathbf{a}$ and $\mathbf{b}$. \hfill$\square$

\subsection{Transport Cost as Alignment Loss}

The OT alignment loss used in the main paper is
\begin{equation}
\label{eq:ot_loss_appendix}
\mathcal{L}_{\mathrm{OT}}
=
\langle \mathbf{P}^\star, \mathbf{C}\rangle.
\end{equation}

This promotes geometric agreement between the 2D latent manifold and the 1D latent sequence.

\subsection{Differentiability and Gradient Computation}

Entropy-regularized OT is smooth in the cost matrix $C$ (Cuturi 2013).
Let $f(C)=\langle P^\star(C),C\rangle$.  
The gradient is
\[
\frac{\partial f}{\partial C}
\;=\;
\mathbf{P}^\star.
\]

Thus, gradients flow from the OT plan into both $\mathbf{X}$ and $\mathbf{Z}$:
\[
\frac{\partial \mathcal{L}_{\mathrm{OT}}}{\partial \mathbf{x}_i}
=
2\sum_{j}
P^\star_{ij}(\mathbf{x}_i-\mathbf{z}_j),
\qquad
\frac{\partial \mathcal{L}_{\mathrm{OT}}}{\partial \mathbf{z}_j}
=
2\sum_{i}
P^\star_{ij}(\mathbf{z}_j-\mathbf{x}_i).
\]

These closed-form expressions ensure efficient backpropagation.

\subsection{Stability and Regularization Properties}

\paragraph{Entropy Regularization.}
The entropic term guarantees:
\begin{itemize}
    \item strictly convex objective,
    \item unique optimal transport plan,
    \item improved numerical stability,
    \item differentiability of the mapping $(C,a,b)\mapsto P^\star$.
\end{itemize}

\paragraph{Sharpness vs. Smoothness.}
As $\varepsilon\to 0$, the plan converges to the classical OT solution.
Larger $\varepsilon$ spreads mass more broadly and stabilizes gradients.

\subsection{Summary}

The 2D to 1D alignment objective is fully characterized by the entropy-regularized
optimal transport problem~\eqref{eq:ot_objective_appendix}, solved uniquely by the
Sinkhorn iterations, and the alignment loss~\eqref{eq:ot_loss_appendix} provides a
stable and fully differentiable measure of geometric consistency between 2D VAE
latents and predicted 1D token sequences.

\section{Additional Details of Implementation.}
\label{appendix:implementation}

This section provides a comprehensive description of the implementation settings used for training the proposed tokenizer. 
\paragraph{General Configuration.}
We fix the global random seed to 42 to ensure reproducibility. Training logs are recorded every 50 iterations, visualizations every 5000 iterations, and checkpoints every 3000 iterations. Mixed-precision training is enabled using bf16. Compiler-level optimizations are disabled (compile = False).
\paragraph{Dataset and Preprocessing.}
All experiments are conducted on the ImageNet. Input images are resized to a spatial resolution of $256\times256$.
\paragraph{Tokenizer and Vector Quantization.}
The temperature parameter is set to $\tau = 0.07$, and EMA updates for codebook vectors are enabled. The tokenizer employs several codebooks each containing 8192 entries, each with embedding dimension 32. The following loss components are used during training:
\begin{itemize}
    \item entropy regularization weight: $0.01$,
    \item VQ quantization loss weight: $0.0$,
    \item commitment loss coefficient: $0.0$,
    \item reconstruction loss weight: $1.0$,
    \item perceptual loss weight: $1.0$ with a warm-up of the first 5000 iterations.
\end{itemize}
Adversarial training is performed with a discriminator weight of 0.2. Diffusion-based augmentation (DiffAug) is enabled. The discriminator incorporates a consistency-regularization term weighted by 4.0 and is activated after 20000 iterations. We employ a DINO-based discriminator using hinge losses for both generator and discriminator; the LeCAM stabilization term is applied with weight 0.001.
\paragraph{Encoder and Decoder Architecture.}
Both encoder and decoder are implemented using Vision Transformers (ViTs):
\begin{itemize}
    \item \textbf{Encoder:} We use the vit\_base\_patch14\_dinov2.lvd142m model with a patch size of $16$, initialized with pretrained weights, and fully fine-tuned during training.
    \item \textbf{Decoder:} We employ the same ViT architecture but without pretrained initialization. The patch size is also $16$, and full tuning is applied. The decoder outputs 128 latent tokens.
\end{itemize}
\paragraph{REPA Representation Alignment.}
Representation alignment is enabled via the REPA module. The REPA backbone is a vit\_large\_patch14\_dinov2.lvd142m transformer with patch size $16$ for global repa and local 1D repa, while the backbone turns to SIGLIP2 and I-JEPA for local 2D repa. The projected feature dimension is set to 1024. The REPA loss is weighted by 0.1, and alignment is performed using the repeat strategy.
\paragraph{Optimization Settings.}
Training is conducted for 50 epochs using the AdamW optimizer with a learning rate of $1\times10^{-4}$. The learning rate follows a cosine decay schedule. The weight decay is 1e-4, and the AdamW momentum parameters are set to $\beta_1 = 0.9$ and $\beta_2 = 0.95$. We apply gradient clipping with a maximum global norm of 1.0. The global batch size is set to 700.
\section{Additional Results for Computational Analysis.}
In this section, we provide further details on the computational analysis of our proposed method.
The base model without the local 2D aligner contains \textbf{250M} parameters, while the local 2D
aligner is scaled across \textbf{100M}, \textbf{200M}, \textbf{400M}, and \textbf{600M} configurations.
The SiT generator includes \textbf{650M} parameters. Therefore, the total number of parameters
involved in the scaling experiment is computed as:
\[
\begin{aligned}
& 250\text{M} \times 4
  + (100\text{M} + 200\text{M} + 400\text{M} + 600\text{M}) \\
& \quad +\, 650\text{M} \times 4 \\
&= 4{,}900\text{M}.
\end{aligned}
\]
In contrast, our pipeline trains the generator only once, which reduces the total parameter
count to:
\[
\begin{aligned}
& 250\text{M} \times 4
  + (100\text{M} + 200\text{M} \\
& \quad + 400\text{M} + 600\text{M})
  + 650\text{M} \\
&= 2{,}950\text{M}.
\end{aligned}
\]
This modification also decreases the total training GFLOPs from \textbf{60K} to \textbf{36.6K},
corresponding to an approximate \textbf{40\%} reduction in computational cost.

Furthermore, training each tokenizer scale requires approximately \textbf{2 days}, and training the
generator requires \textbf{5 days}. Consequently, the total training time is reduced from
\[
(2 \times 4) + (5 \times 4) = 28 \text{ days}
\]
to
\[
(2 \times 4) + 5 = 13 \text{ days},
\]
yielding an approximate \textbf{54\%} reduction in overall training time.

\section{Additional Quantitative Results and Data Scaling Analysis}
In this section, we provide additional quantitative results together with a detailed analysis of data scaling---that is, assessing the behavior of our method when trained on a substantially larger dataset. As shown in~\cref{tab:data_scaling}, incorporating an additional dataset of one million images leads to consistent improvements in both reconstruction and generation quality.
Moreover, our method is compatible with different token capacities; \eg, the 32-token and 64-token configurations in~\cref{tab:token_comparison} exhibit stable performance, demonstrating the flexibility of our tokenizer design. In addition, the scalability of our approach is further validated in~\cref{tab:scaling_performance}, where all reconstruction and generation metrics improve monotonically as the model size increases.
Finally, the impact of our local aligner modules is highlighted in~\cref{tab:tokenizer_metrics}. The introduction of local 1D and 2D aligners leads to substantial gains in reconstruction fidelity, underscoring the importance and effectiveness of the proposed architectural design.

\begin{table}[t]
    \centering
    \caption{Data scaling results on RFID, GFID, PSNR, and SSIM.}
    \label{tab:data_scaling}
    \begin{tabular}{
        l
        @{\hspace{6pt}} c
        @{\hspace{6pt}} c
        @{\hspace{6pt}} c
        @{\hspace{6pt}} c
    }
        \toprule
        \textbf{Setting} & \textbf{RFID} & \textbf{GFID} & \textbf{PSNR} & \textbf{SSIM} \\
        \midrule
        Before Data Scaling & 0.66 & 1.56 & 22.86 & 0.7288 \\
        After Data Scaling  & 0.63 & 1.53 & 22.88 & 0.7289 \\
        \bottomrule
    \end{tabular}
\end{table}

\begin{table}[t]
    \centering
    \caption{Comparison of 32-token and 64-token settings using our method.}
    \label{tab:token_comparison}
    \begin{tabular}{lccc}
        \toprule
        \textbf{Setting} & \textbf{RFID ($\downarrow$)} & \textbf{PSNR ($\uparrow$)} & \textbf{SSIM ($\uparrow$)} \\
        \midrule
        32-token & 0.87 & 20.96 & 0.6993 \\
        64-token & 0.66 & 22.86 & 0.7288 \\
        \bottomrule
    \end{tabular}
\end{table}

\begin{table}[t]
\centering
\setlength{\tabcolsep}{3pt}
\caption{Efficiency comparison of computational overhead at generation time.}
\begin{scriptsize}
\begin{tabular}{lcccc}
\toprule
\textbf{Method} & \textbf{Generator (650M parameters)} & \textbf{gFID} & \textbf{GFLOPs} & \textbf{Latency} \\
\midrule
SoftVQ & SiT(650M) & 1.78 & 7.3K & 7.5ms \\
Ours  & generative component(250M)+SiT(400M) & 1.56 & 7.1K & 5.2ms \\
\bottomrule
\end{tabular}
\end{scriptsize}
\label{tab:fusion_ablation}
\end{table}

\begin{table}[t] 
\centering 
\setlength{\tabcolsep}{4.5pt}  
\caption{\textbf{Ablation study of different fusion strategies.} Comparison between concatenation-based and soft residual fusion under the 400M local 2D aligner and I-JEPA local 1D aligner settings.} \vspace{3pt} 
\begin{small}
\begin{tabular}{lccc} \toprule \textbf{Fusion Strategy} & \textbf{rFID ($\downarrow$)} & \textbf{PSNR ($\uparrow$)} & \textbf{SSIM ($\uparrow$)} \\ \midrule Concatenation Fusion & \textbf{0.66} & \textbf{22.86} & \textbf{0.7288} \\ Soft Residual Fusion & 0.71 & 22.56 & 0.7210 \\ \bottomrule \end{tabular} 
\end{small}
\vspace{-4pt} 
\label{tab:fusion_ablation1122} 
\end{table}

\begin{table}[t]
    \centering
    \caption{Comparison of tokenizer designs. Lower RFID and higher PSNR/SSIM are better.}
    \label{tab:tokenizer_metrics}
    \resizebox{0.90\columnwidth}{!}{
    \begin{tabular}{lccc}
        \toprule
        \textbf{Design} & \textbf{RFID ($\downarrow$)} & \textbf{PSNR ($\uparrow$)} & \textbf{SSIM ($\uparrow$)} \\
        \midrule
        Global                 & 0.85 & 22.09 & 0.7004 \\
        Global + Loc1D         & 0.76 & 22.18 & 0.7020 \\
        Global + Loc1D + Loc2D & 0.70 & 22.25 & 0.7060 \\
        \bottomrule
    \end{tabular}
    }
\end{table}

\begin{table}[t]
    \centering
    \setlength{\tabcolsep}{4.5pt} 
    \renewcommand{\arraystretch}{1.05}
    \caption{Performance of the tokenizer at different model scales. Lower RFID/GFID and higher PSNR/SSIM indicate better performance.}
    \label{tab:scaling_performance}
    \begin{tabular}{lcccc}
        \toprule
        \textbf{Model Scale} & \textbf{RFID ($\downarrow$)} & \textbf{GFID ($\downarrow$)} & \textbf{PSNR ($\uparrow$)} & \textbf{SSIM ($\uparrow$)} \\
        \midrule
        10M  & 0.89 & 1.75 & 22.21 & 0.7103 \\
        100M & 0.83 & 1.74 & 22.34 & 0.7141 \\
        200M & 0.81 & 1.68 & 22.35 & 0.7172 \\
        400M & 0.71 & 1.65 & 22.56 & 0.7211 \\
        600M & 0.66 & 1.56 & 22.86 & 0.7288 \\
        \bottomrule
    \end{tabular}
\end{table}

\section{Additional Qualitative Results.}
In this section, we present additional qualitative results using a broader selection of ImageNet classes. We provide examples from twelve additional categories and illustrate the corresponding generated images. As shown, our method produces high-fidelity and semantically coherent natural images across diverse classes, further demonstrating its strong generative capability.

\begin{figure}[t]
  \centering
  \vspace{-0.5em}
  \includegraphics[width=1.0\linewidth]{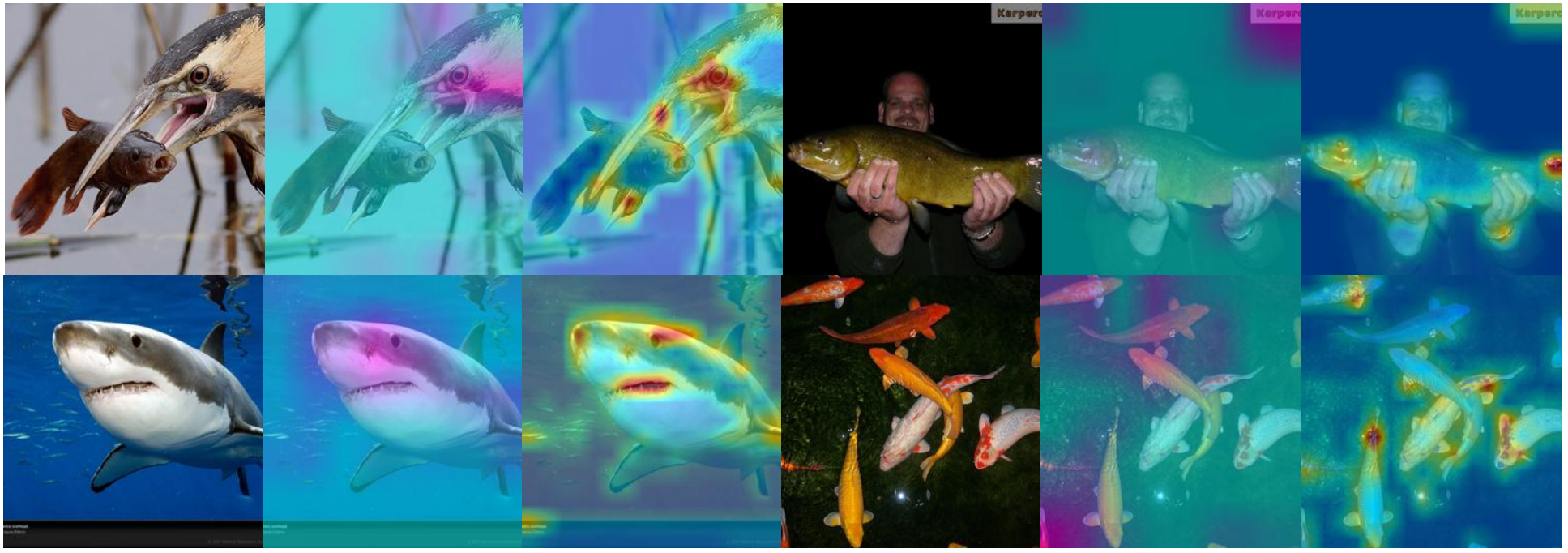}
  \vspace{-0.8em}
  \captionsetup{skip=2pt}
  \caption{Grad-CAM heatmap visualization of token correspondance (\textit{left}: original image, \textit{mid}: global token map, \textit{right}: local 1D+2D token map). Local token groups incorporates more local image details, complementary to the global token group.}
  \vspace{-0.5em}
  \label{fig:oneonecol}
\end{figure}

\begin{figure*}[th!]
    \centering
    \includegraphics[width=0.95\linewidth]{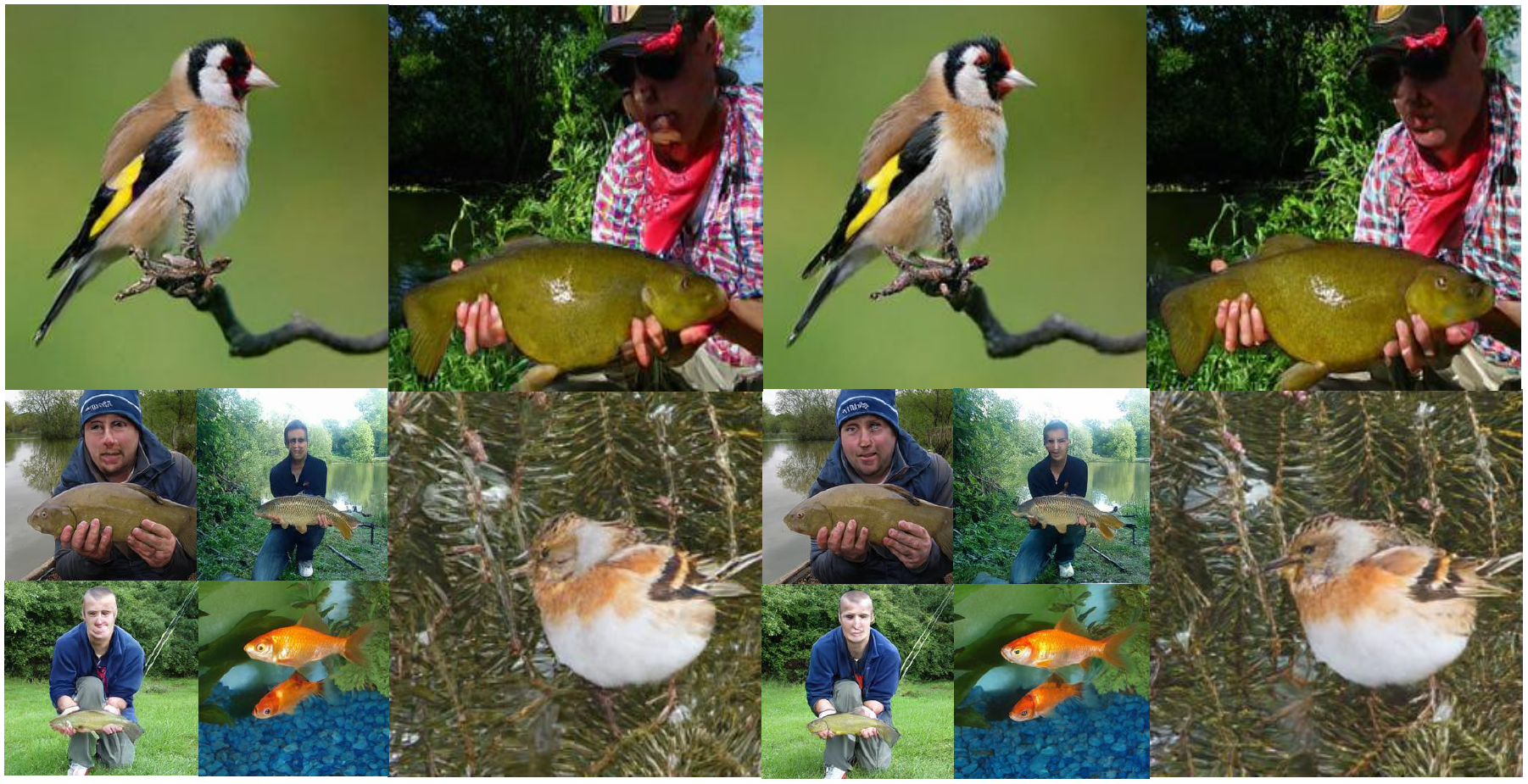}
    \vspace{-0.1in}
    \caption{Qualitative comparison of reconstruction and generation results on ImageNet-1K (256$\times$256) using 64 tokens. We compare models trained with and without efficient scaling of the local aligner. For each image, the left panel shows results without scaling, and the right panel shows results with scaling. Efficient scaling enables the model to recover significantly finer local details, such as the eyes of birds and facial textures of humans, validating the effectiveness of our strategy in improving reconstruction quality.
    }
    \label{fig:reconstruction1122}
    \vspace{-0.2in}
\end{figure*}

\begin{figure*}[th!]
    \centering
    \includegraphics[width=0.95\linewidth]{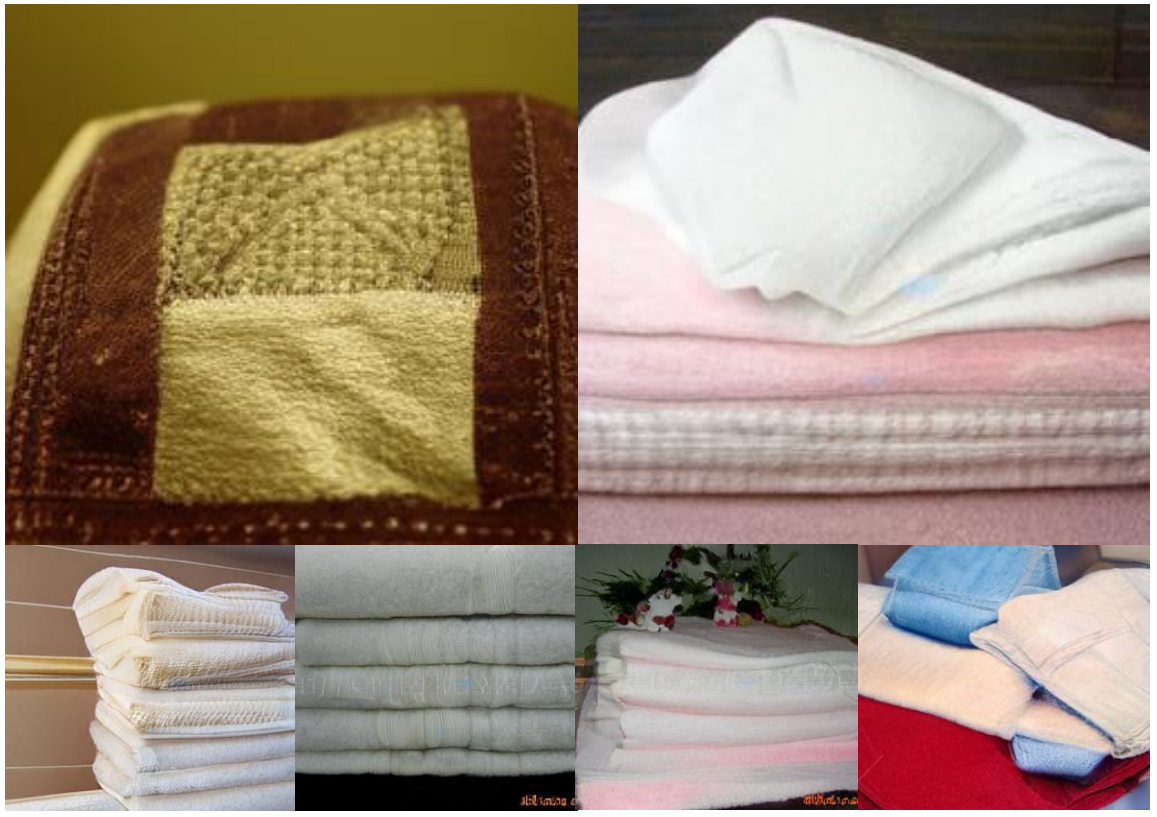}
    \vspace{-0.1in}
    \caption{256$\times$256 generation results produced by SiT-XL/1 using SelfBootTok with 64 tokens. We apply classifier-free guidance (CFG) with a scale of 2.3. The class label used for sampling is ``bath towel'' (class ID 434).
    }
    \label{fig:reconstruction_2}
    \vspace{-0.2in}
\end{figure*}

\begin{figure*}[th!]
    \centering
    \includegraphics[width=0.95\linewidth]{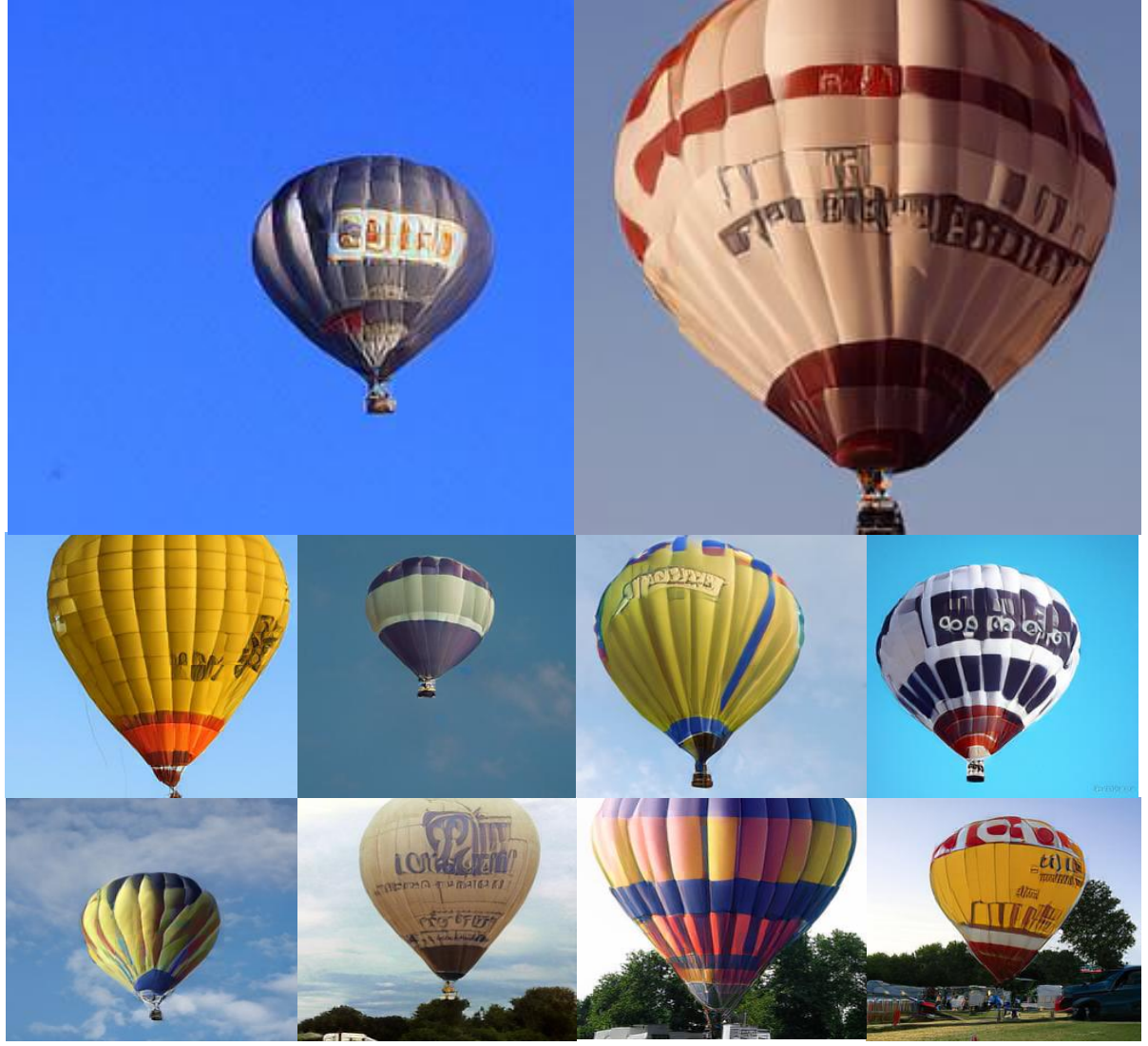}
    \vspace{-0.1in}
    \caption{256$\times$256 generation results produced by SiT-XL/1 using SelfBootTok with 64 tokens. We apply classifier-free guidance (CFG) with a scale of 2.3. The class label used for sampling is ``balloon'' (class ID 417).
    }
    \label{fig:reconstruction_3}
    \vspace{-0.2in}
\end{figure*}

\begin{figure*}[th!]
    \centering
    \includegraphics[width=0.95\linewidth]{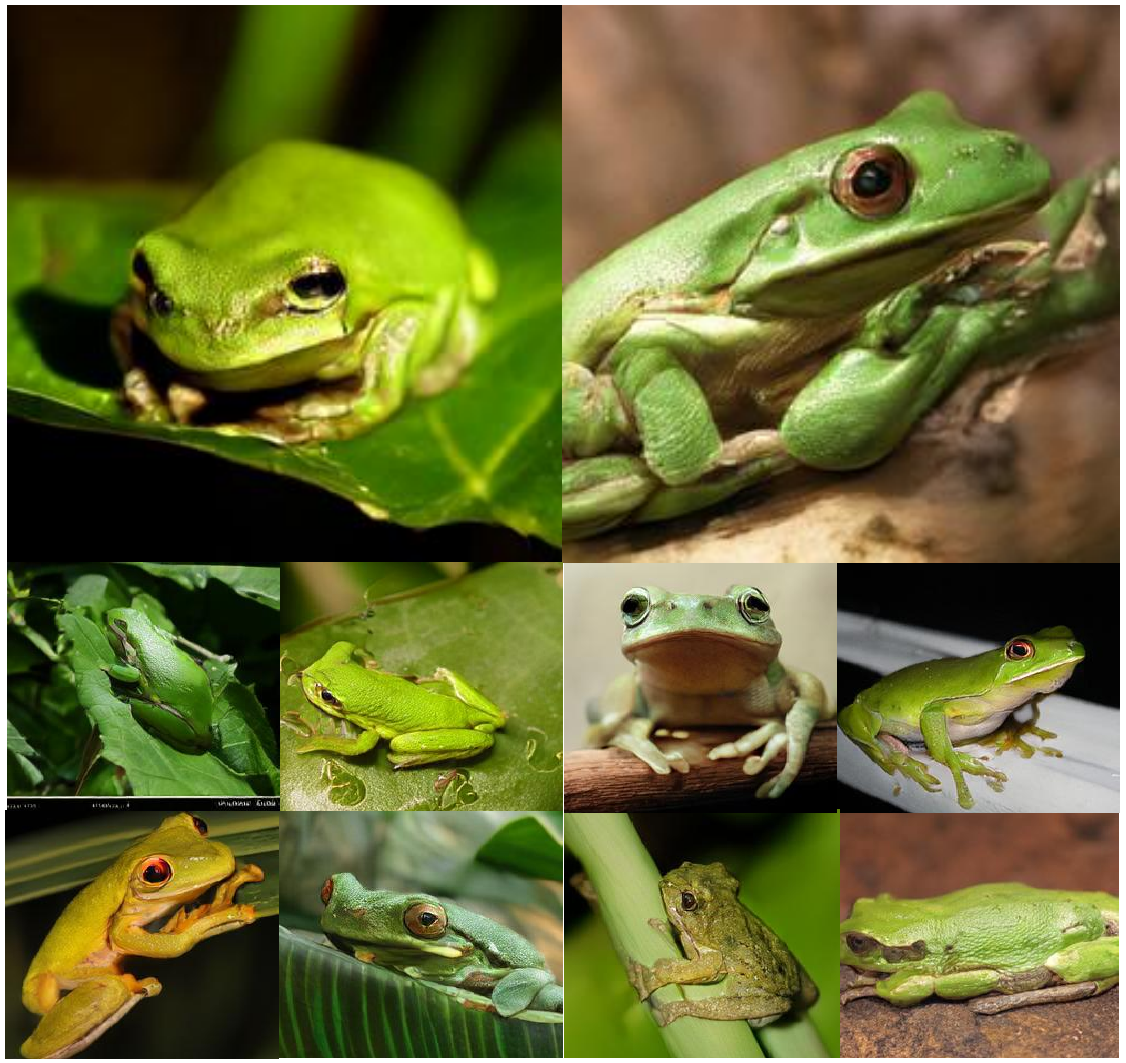}
    \vspace{-0.1in}
    \caption{256$\times$256 generation results produced by SiT-XL/1 using SelfBootTok with 64 tokens. We apply classifier-free guidance (CFG) with a scale of 2.3. The class label used for sampling is ``tree-frog'' (class ID 31).
    }
    \label{fig:reconstruction_4}
    \vspace{-0.2in}
\end{figure*}

\begin{figure*}[th!]
    \centering
    \includegraphics[width=0.95\linewidth]{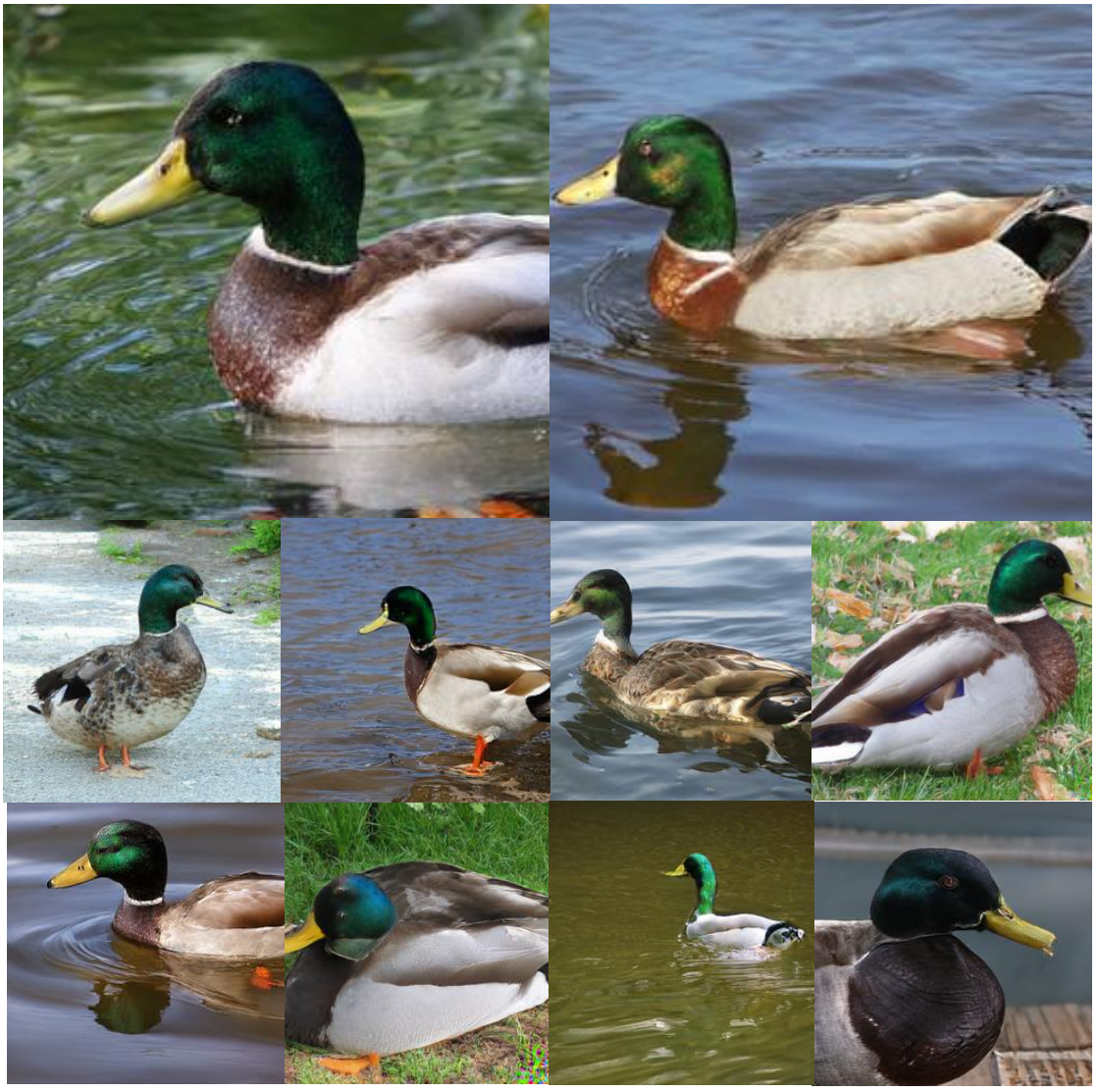}
    \vspace{-0.1in}
    \caption{256$\times$256 generation results produced by SiT-XL/1 using SelfBootTok with 64 tokens. We apply classifier-free guidance (CFG) with a scale of 2.3. The class label used for sampling is ``drake'' (class ID 97).
    }
    \label{fig:reconstruction_5}
    \vspace{-0.2in}
\end{figure*}

\begin{figure*}[th!]
    \centering
    \includegraphics[width=0.95\linewidth]{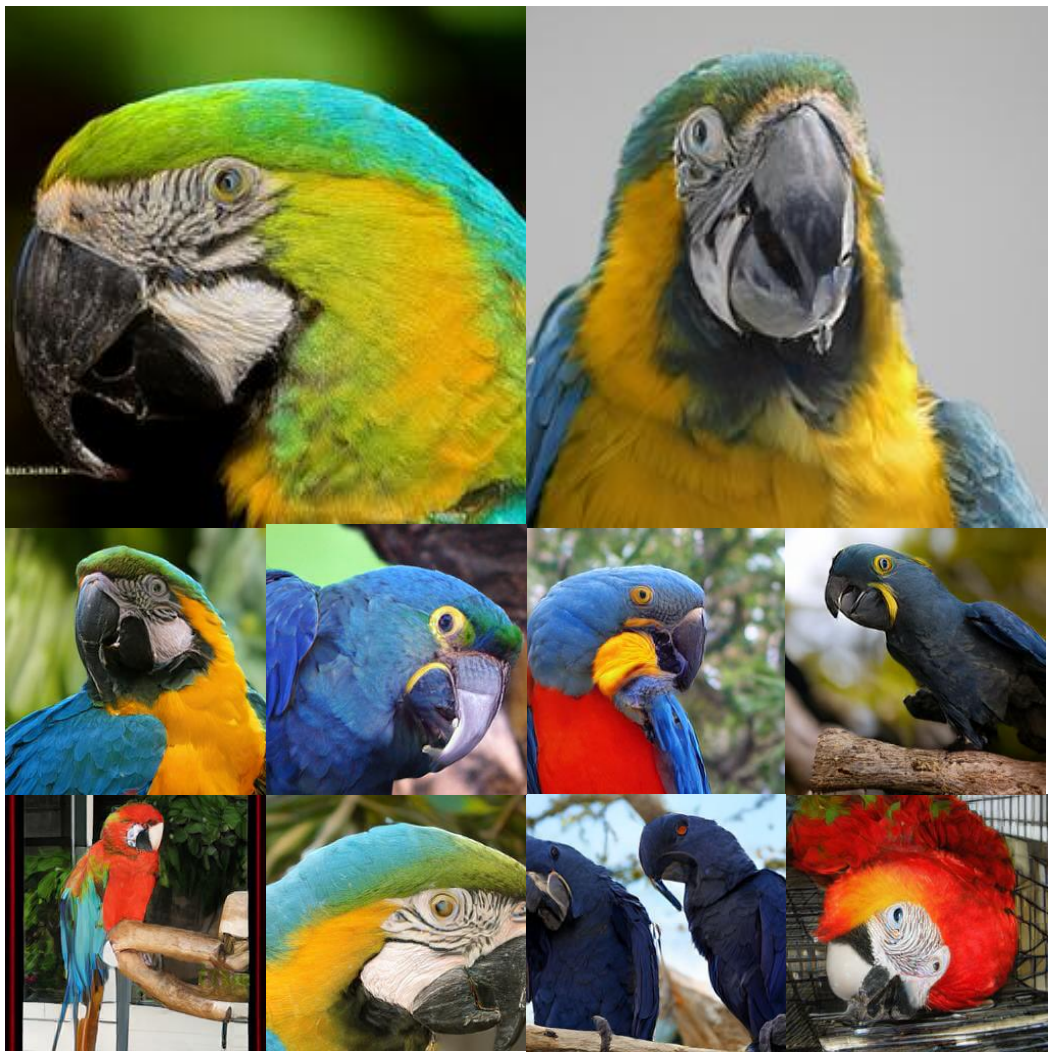}
    \vspace{-0.1in}
    \caption{256$\times$256 generation results produced by SiT-XL/1 using SelfBootTok with 64 tokens. We apply classifier-free guidance (CFG) with a scale of 2.3. The class label used for sampling is ``macaw'' (class ID 88).
    }
    \label{fig:reconstruction_6}
    \vspace{-0.2in}
\end{figure*}

\begin{figure*}[th!]
    \centering
    \includegraphics[width=0.95\linewidth]{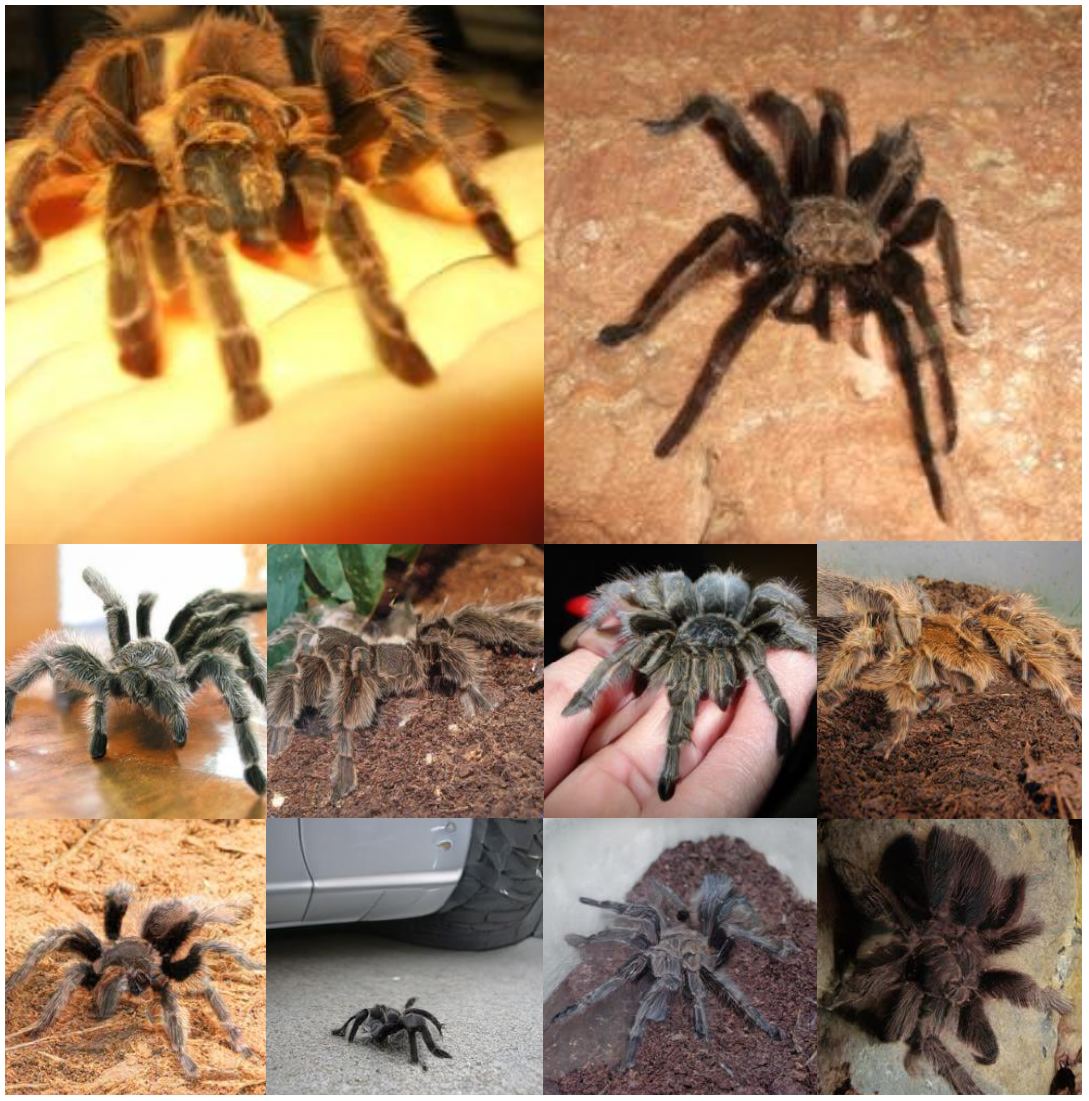}
    \vspace{-0.1in}
    \caption{256$\times$256 generation results produced by SiT-XL/1 using SelfBootTok with 64 tokens. We apply classifier-free guidance (CFG) with a scale of 2.3. The class label used for sampling is ``tarantula'' (class ID 76).
    }
    \label{fig:reconstruction_7}
    \vspace{-0.2in}
\end{figure*}

\begin{figure*}[th!]
    \centering
    \includegraphics[width=0.95\linewidth]{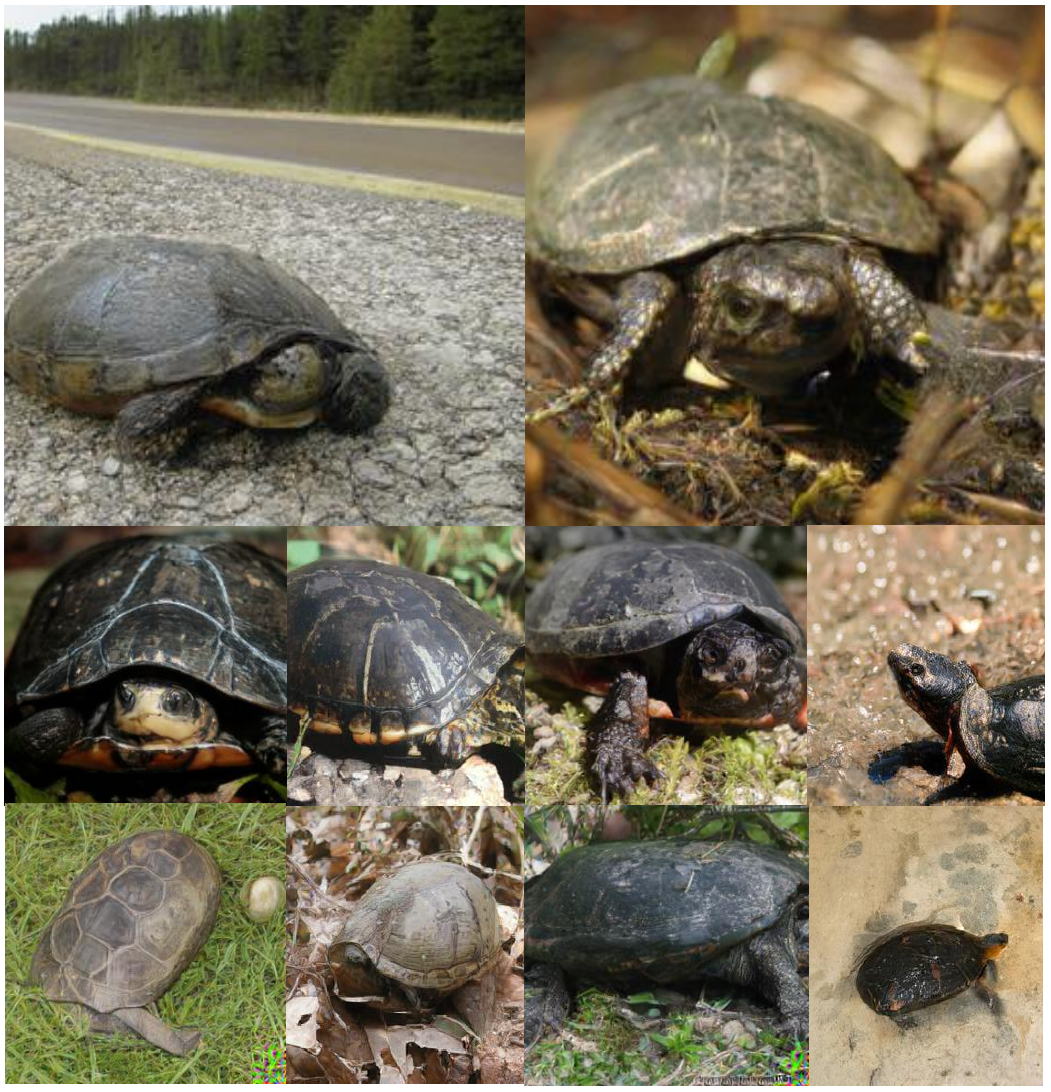}
    \vspace{-0.1in}
    \caption{256$\times$256 generation results produced by SiT-XL/1 using SelfBootTok with 64 tokens. We apply classifier-free guidance (CFG) with a scale of 2.3. The class label used for sampling is ``mud turtle'' (class ID 35).
    }
    \label{fig:reconstruction_8}
    \vspace{-0.2in}
\end{figure*}

\begin{figure*}[th!]
    \centering
    \includegraphics[width=0.95\linewidth]{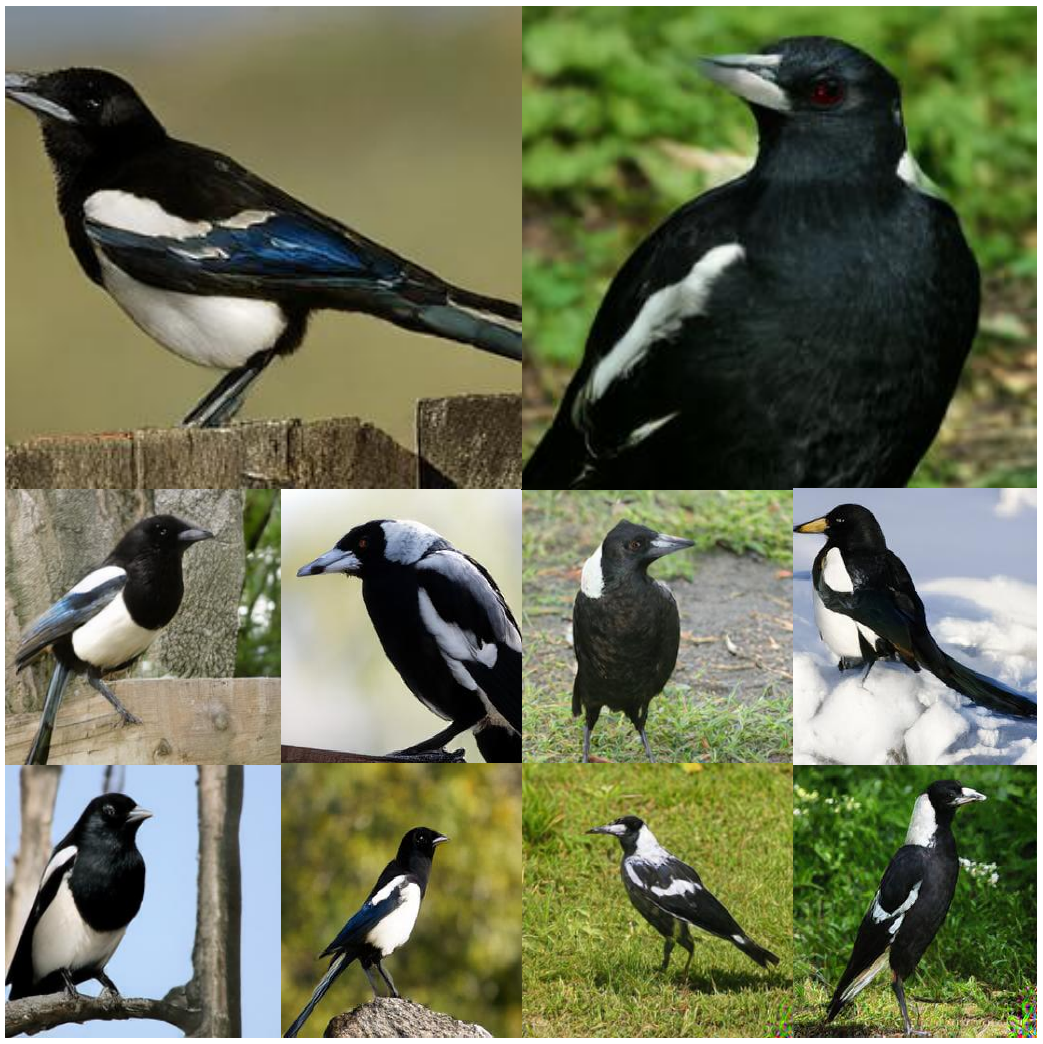}
    \vspace{-0.1in}
    \caption{256$\times$256 generation results produced by SiT-XL/1 using SelfBootTok with 64 tokens. We apply classifier-free guidance (CFG) with a scale of 2.3. The class label used for sampling is ``magpie'' (class ID 18).
    }
    \label{fig:reconstruction_9}
    \vspace{-0.2in}
\end{figure*}

\begin{figure*}[th!]
    \centering
    \includegraphics[width=0.95\linewidth]{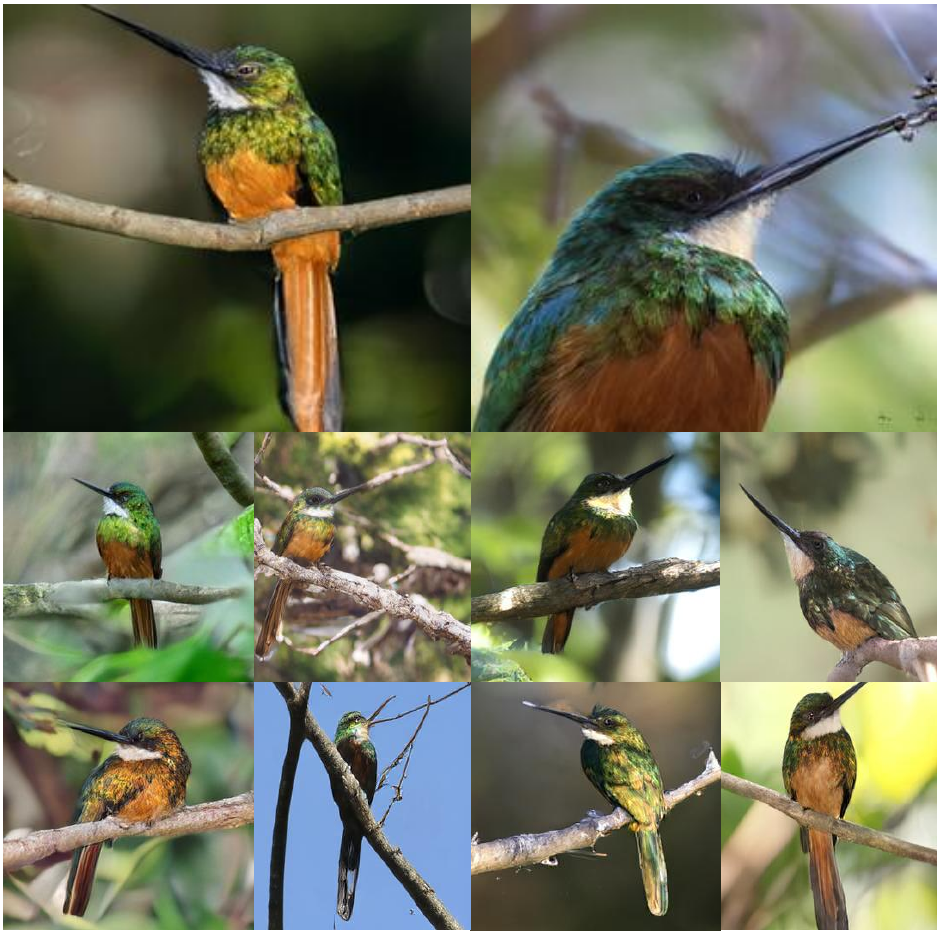}
    \vspace{-0.1in}
    \caption{256$\times$256 generation results produced by SiT-XL/1 using SelfBootTok with 64 tokens. We apply classifier-free guidance (CFG) with a scale of 2.3. The class label used for sampling is ``jacamar'' (class ID 95).
    }
    \label{fig:reconstruction_10}
    \vspace{-0.2in}
\end{figure*}

\begin{figure*}[th!]
    \centering
    \includegraphics[width=0.95\linewidth]{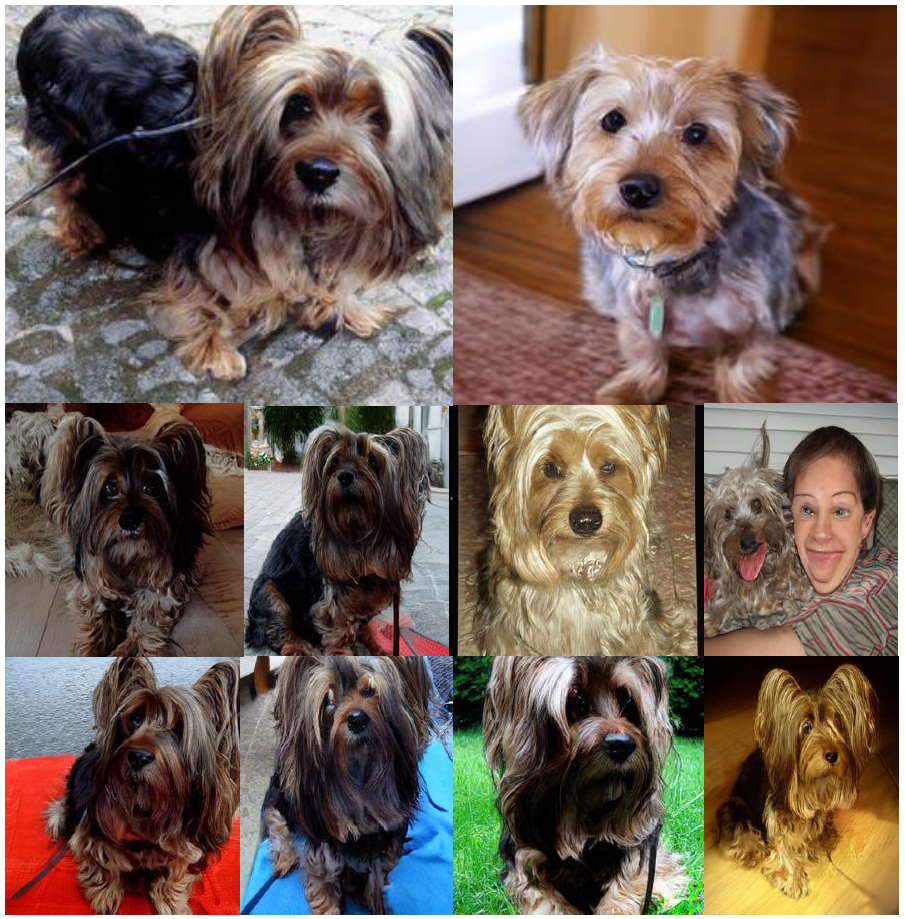}
    \vspace{-0.1in}
    \caption{256$\times$256 generation results produced by SiT-XL/1 using SelfBootTok with 64 tokens. We apply classifier-free guidance (CFG) with a scale of 2.3. The class label used for sampling is ``silky terrier, Sydney silky'' (class ID 201).
    }
    \label{fig:reconstruction_11}
    \vspace{-0.2in}
\end{figure*}

\begin{figure*}[th!]
    \centering
    \includegraphics[width=0.95\linewidth]{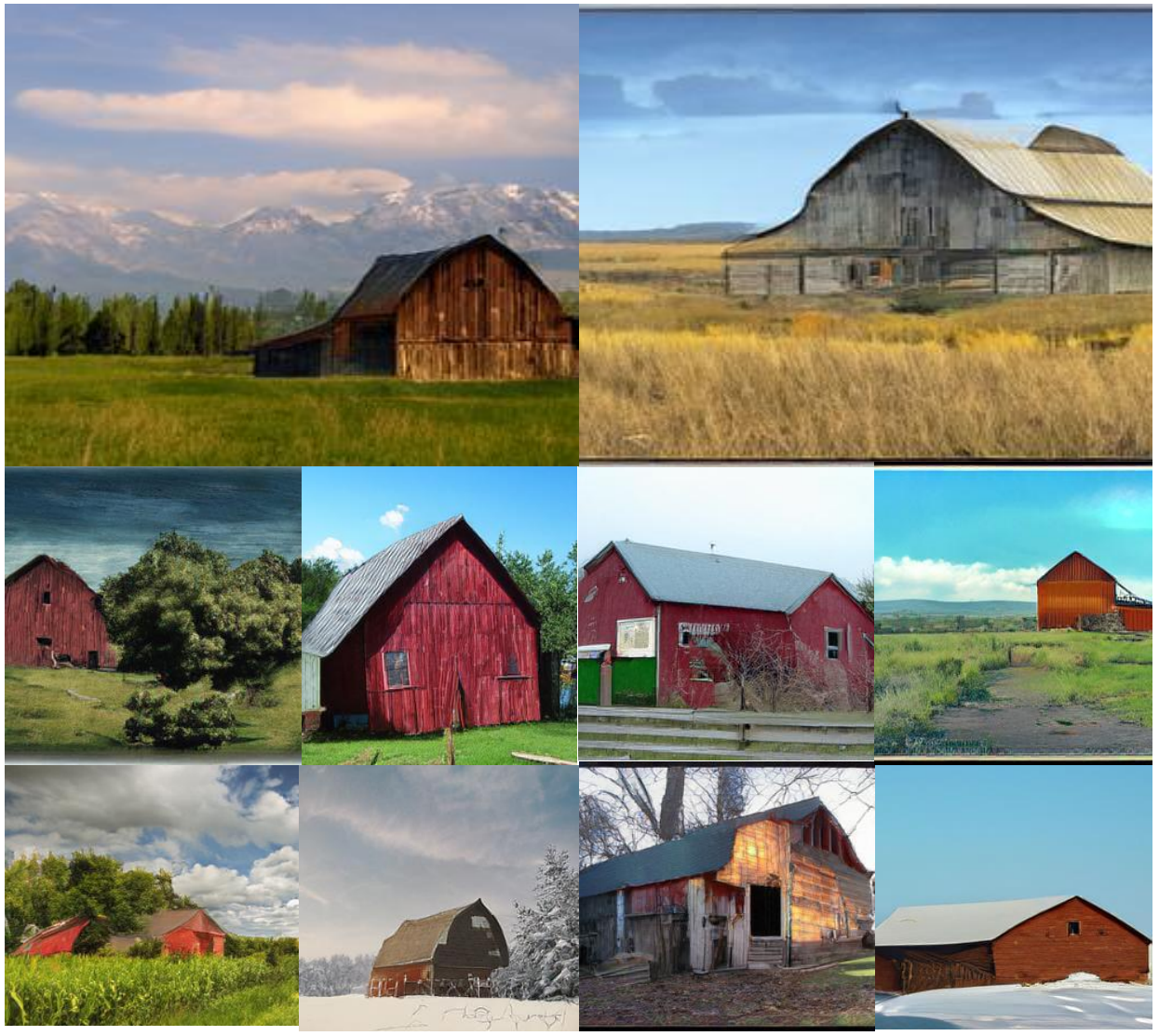}
    \vspace{-0.1in}
    \caption{256$\times$256 generation results produced by SiT-XL/1 using SelfBootTok with 64 tokens. We apply classifier-free guidance (CFG) with a scale of 2.3. The class label used for sampling is ``barn'' (class ID 425).
    }
    \label{fig:reconstruction_12}
    \vspace{-0.2in}
\end{figure*}

\begin{figure*}[th!]
    \centering
    \includegraphics[width=0.95\linewidth]{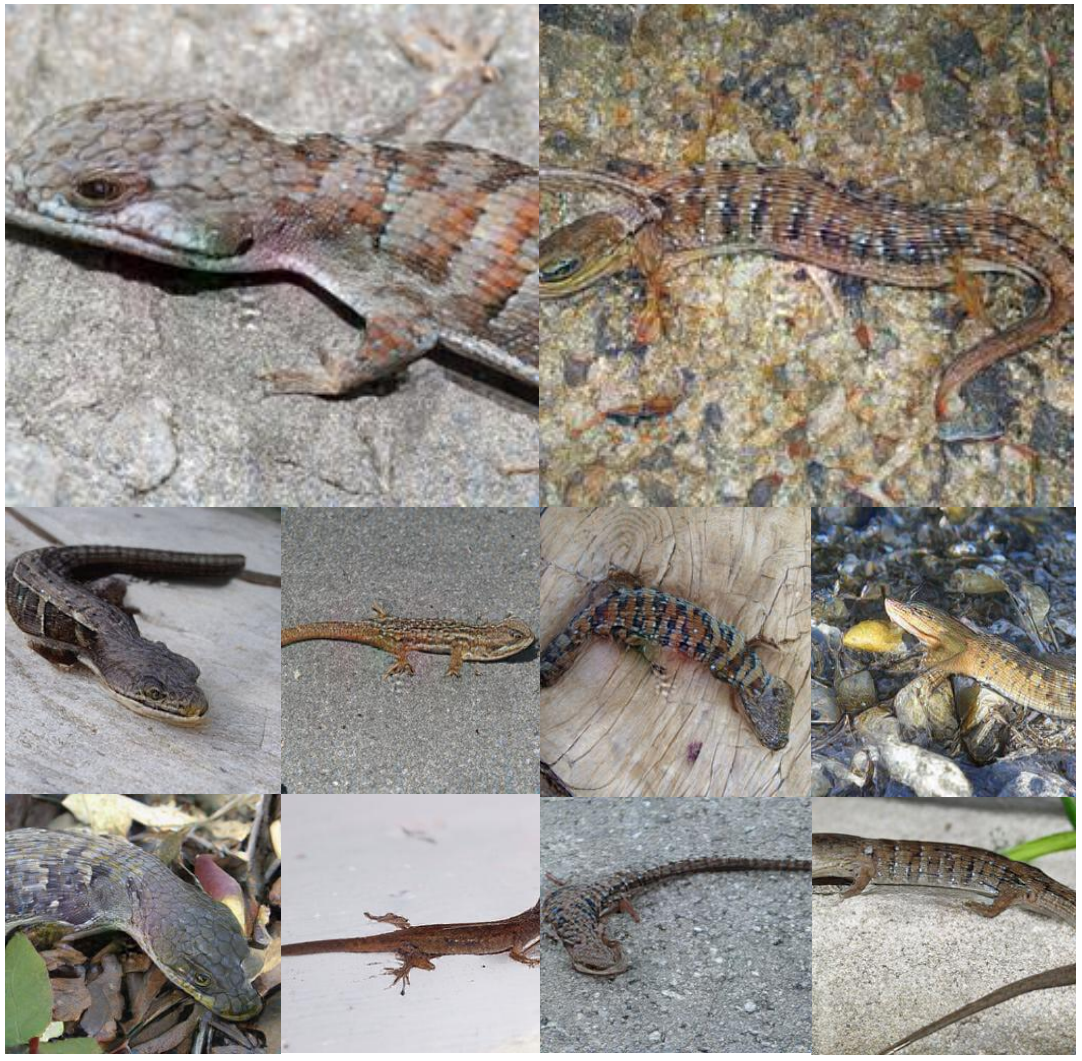}
    \vspace{-0.1in}
    \caption{256$\times$256 generation results produced by SiT-XL/1 using SelfBootTok with 64 tokens. We apply classifier-free guidance (CFG) with a scale of 2.3. The class label used for sampling is ``alligator lizard'' (class ID 44).
    }
    \label{fig:reconstruction_13}
    \vspace{-0.2in}
\end{figure*}

\section{Appendix References}
\begingroup
\renewcommand{\section}[2]{}%

\endgroup

\end{document}